\documentclass{article} % For LaTeX2e
\usepackage{iclr2026_conference,times}

\usepackage{amsmath,amsfonts,bm}

\def\eqref#1{equation~\ref{#1}}
\def\1{\bm{1}}

\DeclareMathAlphabet{\mathsfit}{\encodingdefault}{\sfdefault}{m}{sl}
\SetMathAlphabet{\mathsfit}{bold}{\encodingdefault}{\sfdefault}{bx}{n}

\usepackage{hyperref}
\usepackage{url}

\usepackage[utf8]{inputenc} % allow utf-8 input
\usepackage[T1]{fontenc}    % use 8-bit T1 fonts
\usepackage{hyperref}       % hyperlinks
\usepackage{url}            % simple URL typesetting
\usepackage{booktabs}       % professional-quality tables
\usepackage{amsfonts}   
\usepackage{nicefrac}       % compact symbols for 1/2, etc.
\usepackage{microtype}      % microtypography
\usepackage{xcolor}         % colors

\usepackage{lineno}
\usepackage{graphicx}
\usepackage{tikz}
\usepackage{amsmath}
\usepackage{amssymb}
\usepackage{bbm}
\usepackage{adjustbox}
\usepackage{multirow}
\usepackage{wrapfig}
\usepackage{tabularx}

\definecolor{darkblue}{rgb}{0, 0, 0.5}
\hypersetup{colorlinks=true, citecolor=darkblue, linkcolor=darkblue, urlcolor=darkblue}

 \title{Tversky Neural Networks: Psychologically Plausible Deep Learning with 
 Differentiable Tversky Similarity}

\author{%
  Moussa Koulako Bala Doumbouya$\quad$
  Dan Jurafsky$\quad$
  Christopher D. Manning\\
  \\
Department of Computer Science, 
353 Jane Stanford Way; Stanford, CA 94305\\
\texttt{\{moussa, jurafsky, manning\}@stanford.edu}\\
}

\iclrfinalcopy % Uncomment for camera-ready version, but NOT for submission.

\begin{document}

\maketitle

\lhead{Preprint} % for Arxiv

\iffalse
\begin{abstract}
Work in psychology has highlighted that the geometric model of similarity standard in deep learning is not psychologically plausible because its metric properties, such as symmetry, do not align with human judgment of similarity.
%
In contrast, \citet{tversky1977features} proposed a psychologically plausible theory of similarity representing objects as sets of features and their similarity as linear combinations of measures of their common and distinctive features.
%
However, this model has not been used in deep learning, in part because of the challenge of incorporating discrete set operations into gradient based machine learning.
%
Here we develop a differentiable parameterization of Tversky's similarity that is learnable through gradient descent, and show
that it is a beneficial replacement for neural modules that employ geometric similarity such as the linear projection layer.
%
For example,
unlike linear projection layers, a single Tversky Projection layer can model the {\sc xor} function,
%
and replacing linear projection layers with a Tversky projection layers decreases the perplexity of GPT-2 trained on PTB by 7.5\%, and its parameter count by 34.8\%.
%
Our qualitative analysis demonstrates that Tversky Projection layers are also more interpretable than the linear projection layers they replace.
Our work offers a new paradigm for thinking about the  similarity model implicit in modern deep learning.
\end{abstract}
\fi

\begin{abstract}
    Work in psychology has highlighted that the geometric model of similarity standard in deep learning is not psychologically plausible because its metric properties such as symmetry do not align with human perception of similarity.
    In contrast, \citet{tversky1977features} proposed an axiomatic theory of similarity with psychological plausibility based on a representation of objects as sets of features, and their similarity as a function of their common and distinctive features.
    This model of similarity has not been used in deep learning before, in part because of the challenge of incorporating discrete set operations. 
    In this paper, we develop a differentiable parameterization of Tversky's similarity that is learnable through gradient descent, and derive basic neural network building blocks such as the \emph{Tversky projection layer}, which unlike the linear projection layer can model non-linear functions such as  {\sc xor}.
    Through experiments with image recognition and language modeling neural networks, we show that the Tversky projection layer is a beneficial replacement for the linear projection layer.
    For instance, on the NABirds image classification task, a frozen ResNet-50 adapted with a Tversky projection layer achieves a 24.7\% relative accuracy improvement over the linear layer adapter baseline.
    With Tversky projection layers, GPT-2's perplexity on PTB decreases by 7.8\%, and its parameter count by 34.8\%.
    Finally, we propose a unified interpretation of both types of projection layers as computing similarities of input stimuli to learned prototypes for which we also propose a novel visualization technique highlighting the interpretability of Tversky projection layers. 
    Our work offers a new paradigm for thinking about the similarity model implicit in modern deep learning, and designing neural networks that are interpretable under an established theory of psychological similarity.
\end{abstract}

\section{Introduction}

The geometric model of similarity is ubiquitously used in modern neural networks. Their architectures include operations that assess the similarity between objects using metric
similarity functions such as the vector dot product or cosine similarity.
For instance, linear projection layers, also known as \emph{dense}, \emph{feed forward}, or \emph{fully connected}, output the dot-product similarity of the input vector with the columns of its weight matrix, which are concept prototypes.
Classification modules typically normalize these similarities with {\sc softmax} to form valid probability distributions \citep{bridle1990probabilistic,lecun1998gradient}.
Convolution layers of image recognition neural networks compute the dot product of image regions and convolution kernels, which represent prototypical patterns.
LSTM gates compute dot product similarities of combinations of inputs, hidden states, and cell states with prototypical \emph{input}, \emph{forget}, and \emph{output} gating patterns \citep{hochreiter1997long,graves2005bidirectional,graves2013generating}.
Semantic word embedding models such as word2vec \citep{mikolov2013efficient} and GloVe \citep{pennington2014glove} compute the dot-product and cosine similarity of word representations with prototypical embeddings to predict words.
Neural language models apply the same classification mechanism to output discrete words or subwords at each time step \citep{bengio2000neural, sutskever2014sequence,devlin2018bert, radford2018improving, achiam2023gpt}.
Attention weights \citep{bahdanau2014neural, vaswani2017attention} are the normalized dot-product similarity of \emph{queries} and \emph{keys}, which are themselves obtained using linear projection layers.
% GEOMETRIC MODEL IN LOSS FUNCTIONS
Geometric similarity representations have also been widely used in machine learning objective function design for over 200 years, from least squares methods \citep{legendre1806nouvelles} to modern approaches including L2 loss \citep{hadsell2006dimensionality}, 
%Triplet Loss \citep{schroff2015facenet}, 
Large Margin Cosine Loss \citep{wang2018cosface},
%Additive Angular Margin Loss\citep{deng2019arcface}, 
and 
Noise Contrastive Estimation objective \citep{oord2018representation, chen2020simple}. These objectives span both supervised and self-supervised learning paradigms.
%
%Neural networks employing the geometric model of similarity in their architecture or training objectives have been widely successful, resulting in applications such as modern large language models, which have large positive and negative societal impacts \cite{bommasani2021opportunities}.

However, \citet{tversky1977features} famously challenged this geometric representation of similarity, showing its psychological implausibility due to its fundamental inability to model phenomena such as humans' asymmetric judgment of similarity.
For instance, ``We say that \emph{`the son resembles the father'} rather than \emph{`the father resembles the son'}.''
He addressed this problem with his axiomatic theory based on a feature matching process in which objects are represented as sets of features and their similarity is measured as a linear combination of measures of their common and distinctive features.
Tversky's similarity has been shown to model human judgment more accurately than metric similarity measures in various empirical studies \citep{gati1984weighting,ritov1990differential,siegel1982experimental,rorissa2007relationships}.
Moreover, over the past decade, several works have highlighted the limitations of the geometric model of similarity and related parallelogram model of semantic relations \citep{Garg_Enright_Madden_2015,Chen_Peterson_Griffiths_2017,peterson2020parallelograms,zhou2022problems}, and advocated for more psychologically principled similarity measures such as Tversky's model, which has strong empirical support.

Nevertheless, a differentiable expression of Tversky's similarity function has not been proposed to date, which prevents its incorporation into neural networks trained using gradient-based methods
\citep{rumelhart1986learning,lecun1998gradient}.
% \citep{rumelhart1986learning,lecun1998gradient,pascanu2013difficulty,dean2012large,kingma2014adam,hinton2012rmsprop,duchi2011adaptive,shazeer2018adafactor}.
The formulation of a differentiable Tversky similarity function is non-trivial because it employs measures of set intersections and differences, which are not differentiable with respect to object features comprising those sets.
To address this gap, we propose a novel representation of features as vectors of the same dimensionality as object vectors, and a dual representation of objects both as vectors and as sets, such that an object is the set of features with which it has a positive dot product. This representation of objects and features enables us to define differentiable measures of set intersections and differences, which we use to construct a differentiable Tversky similarity function suitable for deep learning. We also propose the Tversky Projection neural layer, which is analogous to the linear projection layer but is based on Tversky's model of similarity.

We introduce our proposed differentiable similarity function, and basic neural network modules, along with a novel method for interpreting and visualizing deep projection layers in Section~\ref{sec:methods}.
%
% Section~\ref{sec:quantitative-experiments} and Section~\ref{sec:quantitative-analysis} present our quantitative and qualitative results showing the benefits of using Tversky Projection layers.
%
Section~\ref{sec:quantitative-experiments} demonstrates the effectiveness of Tversky Projection layers in state-of-the-art architectures, including ResNet-50 and GPT-2, on image recognition and language modeling tasks. Our experiments show that the use of Tversky projection layers can lead to significant improvements in both accuracy and parameter efficiency.
In particular, we show that the use of a Tversky Projection layer instead of the final linear projection layer in ResNet-50 can result in up to 24\% accuracy improvement in image recognition tasks.
We also show that replacing linear projection layers in the language modeling head, and attention blocks in GPT-2 can result in a 7.5\% reduction in perplexity and 34.8\% reduction in parameter count.
Section~\ref{sec:results:qualitative-analysis} provides various qualitative analyses of Tversky projection layers, including results highlighting their principled explanability under Tversky's theory of similarity.
Overall the main contribution of this work is to provide a foundation for 
efficient and interpretable neural networks based on an established theory of psychological similarity.

%We also propose a Tversky variant of the linear projection layer and the attention layer, which are ubiquitously used in state-of-the-art neural networks.
%
%Unlike the linear projection layer, a single Tversky Projection layer can represent non-linear functions such as the xor function.
%A single multi-headed Tversky attention layer can perform sequence reordering with higher accuracy than baseline, and doesn't employ the `query'-`key' mechanism, which was likely introduced to enable asymmetric attention matrices.

%To show the psychological plausibility of our proposed similarity function, we inspect whole word representations learned by our proposed Tversky word2vec, which enables algebraic manipulations of concepts as sets of features, permitting the extraction of semantic relations not possible with prior approaches such as the parallelogram model.

%We first  show that unlike the linear projection layer, a single Tversky Projection layer can represent non-linear functions such as the {\sc xor} function.  We then
%apply our neural modules in large-scale machine learning experiments on image recognition and language modeling,  showing that Tversky Neural Networks perform better than their feedforward counterparts, with the added benefit of being interpretable, parameter efficient and psychologically plausible.

%\section{Tversky's Similarity}
%\label{sec:tversky}
%\input{sec_related_work}

\section{Methods}
\label{sec:methods}

\subsection{Tversky's Similarity}
% See also: https://pigeon.psy.tufts.edu/avc/dblough/theory.htm
Tversky's model of similarity \citep{tversky1977features} has emerged as an influential theoretical representation of human similarity judgment, supported by extensive empirical evidence from cognitive psychology \citep{tversky1982similarity,gati1984weighting,goldstone1994role,medin1993respects,rorissa2007relationships}. His work challenged the geometric model of similarity by demonstrating that humans systematically violate metric axioms such as minimality, symmetry, and triangle inequality when assessing similarity, and proposed a theoretical framework in which objects are represented as sets of features and their similarity assessment as a feature matching process.
Formally, Tversky's asymmetric model of similarity of object $a$ to object $b$ defined as feature sets $A$ and $B$ is a function $F$ of their common and distinctive features. Tversky showed in his axiomatic framework that $F$ is a linear combination of measures of its set parameters, namely: the common features of $a$ and $b$, the distinctive features of $a$, and the distinctive features of $b$ (Equation~\ref{eq:tversky-constrast-model}). 
In Section~\ref{sec:differentiable-tversky-similarity}, we introduce a differentiable parameterization of this function, making it suitable for gradient-based machine learning.
\begin{align}
S(a, b) = F(A\cap B, A-B, B-A)
%\label{eq:tversky-model}\\
%S(a, b) 
= \theta f(A\cap B) - \alpha f(A-B) - \beta f(B-A)
\label{eq:tversky-constrast-model}
\end{align}

\subsection{Differentiable Tversky Similarity}
\label{sec:differentiable-tversky-similarity}
This section presents our proposed differentiable parameterization of Tversky's similarity function, which is constructed with 
a representation of features as vectors of the same dimensionality as objects, and
a dual representation of objects as vectors and as sets.

\textbf{Dual Representation of Objects as Vectors and as Sets:}
Given the learnable finite universe $\Omega$ of features vectors $f_k \in \mathbb{R}^d$, and an object represented as the vector $x \in \mathbb{R}^d$,
we propose $x \cdot f_k$ to be the scalar measure of feature $f_k$ in $x$, and a second representation of $x$ as the set
$X=\{f_k \in \Omega \vert x \cdot f_k>0\}$ of features with which $x$ has a positive dot product.

\textbf{Salience:}
Tversky hypothesized that the relative salience of stimuli, or prominence of their features, determines the direction of asymmetry in human's judgment of similarity. The less salient stimulus (e.g. the son) is assessed to be more similar to the more salient stimulus (e.g. the father) than vice versa.
Following Tversky's theory and our proposed representation, the salience of features in an object $A$, which is the sum of the measures of all features present in the object is $f(A)$ (Equation~\ref{eq:saliency}).
\begin{equation}
f(A) = \sum\nolimits_{k=1}^{|\Omega|} a \cdot f_k \cdot \mathbbm{1}[a \cdot f_k >0]
\label{eq:saliency}
\end{equation}

\textbf{Feature Set Intersections:}
To measure the common features of objects $A$ and $B$, $f(A \cap B)$ (Equation~\ref{eq:intersection-measure}), we propose a function $\Psi$ to aggregate measures of the features present in both $a$ and $b$, and experiment with values \emph{min}, \emph{max}, \emph{product}, \emph{mean}, \emph{gmean} and \emph{softmin}.
This function corresponds to the 
\emph{intersection reduction} hyperparameter of Tversky neural modules.
% We experimented with \emph{products}, \emph{mean}, \emph{min}, \emph{max}, \emph{softmin}, and \emph{gmean}.
\begin{align}
f(A \cap B) &=  
 \sum\nolimits_{k=1}^{|\Omega|} \Psi \bigl( a \cdot f_k, b \cdot f_k \bigr)
 \times \mathbbm{1}\bigl[ a \cdot f_k > 0 
 \land b \cdot f_k > 0 \bigr]
 \label{eq:intersection-measure}
\end{align}

\textbf{Feature Set Difference:}
$f(A-B)$ is a measure of features present in $A$ but not present in $B$ (Equation~\ref{eq:difference-measure-ignore-match}). We propose an alternate form of this measure that accounts for features that are present in both $A$ and $B$, but in greater amount in $A$ (Equation~\ref{eq:difference-measure-substract-match}). These two measures of set difference respectively correspond to the values \emph{ignorematch} and \emph{substractmatch} of the \emph{difference reduction} hyperparameter of Tversky neural modules.
\begin{align}
f^i(A - B) &= 
    \sum\nolimits_{k=1}^{|\Omega|} \bigl( a \cdot f_k \bigr)
    \times \mathbbm{1}\bigl[ a \cdot f_k > 0 
    \land b \cdot f_k \leq 0 \bigr] 
    \label{eq:difference-measure-ignore-match}
\\
f^s(A - B) &= f^i(A - B) + 
    \sum\nolimits_{k=1}^{|\Omega|} \bigl( a \cdot f_k - b \cdot f_k \bigr)
    \times \mathbbm{1}\bigl[ 
    b \cdot f_k > 0 
    \land 
    a \cdot f_k > b \cdot f_k \bigr] 
\label{eq:difference-measure-substract-match}
\end{align}

\subsection{Tversky Neural Network Modules}
We propose two basic building blocks for Tversky neural networks, the \emph{Tversky Similarity Layer}, which is analogous to metric similarity functions such as dot product or cosine similarity, and the \emph{Tversky Projection Layer}, analogous to the  contemporary \emph{fully connected layer}.
\paragraph{Tversky Similarity Layer:}
This layer, formalized in Equation~\ref{eq:tversky-similarity-layer}, calculates the similarity of  object $a \in \mathbb{R}^d$ to object $b \in \mathbb{R}^d$. Its learnable parameters are a feature bank $\Omega$, and the $\alpha$, $\beta$ and $\theta$ scalar parameters of Tverky's contrast model of similarity (Equation~\ref{eq:tversky-constrast-model}).
\begin{align}
        \mathcal{S}^{\Omega,\alpha,\beta,\theta}(a,b) \colon
\begin{cases}
  \mathbb{R}^d\times \mathbb{R}^d & \longrightarrow\,\,\, \mathbb{R} \\
  (a, b)          & \longmapsto\,\,\, 
\begin{bmatrix}
\theta \\
-\alpha \\
-\beta \\
\end{bmatrix} 
\cdot
\begin{bmatrix}
f(A\cap B) \\
f(A-B) \\
f(B-A) \\
\end{bmatrix}
\end{cases}
\label{eq:tversky-similarity-layer}
\end{align}

\paragraph{Tversky Projection Layer:}
This non-linear projection (Equation~\ref{eq:tversky-projection-layer}), calculates the similarity of its input $a \in \mathbb{R}^d$ to each prototype in the ordered set of $p$ prototypes $\Pi_i \in \mathbb{R}^d$, yielding a vector in $\mathbb{R}^p$. 
% Figure~\ref{fig:constructed-tversy-xor} illustrates a constructed Tversky projection layer that models the $xor$ function along with its parameters and their set-based interpretation.
% Our results in Section~\ref{sec:quantitative-experiments} and \ref{sec:quantitative-analysis} show that this layer could be a valuable replacement of the state of the art fully connected layer.
\begin{align}\mathcal{P}^{\Omega,\alpha,\beta,\theta, \Pi}(a) \colon
\begin{cases}
  \mathbb{R}^d  & \longrightarrow\,\,\, \mathbb{R}^p \\
  a          & \longmapsto\,\,\,
    \begin{bmatrix}
        \mathcal{S}^{\Omega,\alpha,\beta,\theta}(a,\Pi_0) \\
        \mathcal{S}^{\Omega,\alpha,\beta,\theta}(a,\Pi_1) \\
        \vdots \\
        \mathcal{S}^{\Omega,\alpha,\beta,\theta}(a,\Pi_{p-1}) \\
    \end{bmatrix}
\end{cases}
\label{eq:tversky-projection-layer}
\end{align}

\subsection{Interpretation of Projection Layers:}
Both the linear and Tversky projection layers output vectors in which each dimension is the similarity of the input to a prototype; however, they differ in the employed similarity function.
Linear projection layers
 compute dot-product similarities. This is an application of the geometric model of simlarity, in which the features of objects and prototypes are scalar values arranged in a cartesian coordinate space and similarity is measured as the oriented length of the object vector's projection onto the prototype vector. Dot product similarities have metric properties such as asymmetry and triangle inequality that have been empirically shown to not align with human perception, and proven to be incapable of modeling asymmetric relations.
Tversky projection layers
compute Tversky similarity with learnable parameters  prototype vectors $\Pi$, feature vectors $\Omega$, and Tversky's contrast model's weights $\alpha, \beta, and \theta$. The number of features $|\Omega|$ can be varied without affecting the output dimensionality $p=|\Pi|$.

% Figure~\ref{fig:xornet-with-various-features} shows the parameters and decision boundaries of a Tversky projection layer modeling the $xor$ function with various feature counts $|\Omega|$.

\subsubsection{Tversky Feature Sharing}
Tversky feature banks and prototype banks can be shared across various layers in a neural network in semantically justifiable ways.
For instance, in a Tversky GPT-2, 
%the attention token projection layers, 
the output projection layers in attention blocks, and the final language modeling head can all share the same feature bank $\Omega$ as this parameter is semantically compatible across all those layers: it represents token features.
%
% The Tversky similarity layer used to compute attention weights across all layers can share the same attention features $\Omega^a$ because these represent attention token features.
%
Tversky projection layers that are language modeling heads can also use token embeddings as token prototypes. This parameter sharing strategy is similar to \emph{weight tying} widely used in language models employing a linear projection layer to classify output tokens \citep{inan2016tying}.
Multiple Tversky similarity or projection layers can share their feature or prototype bank if semantically compatible. They can also share the same feature bank while maintaining separate prototypes, or keep all their parameters separate.
Our results in Section~\ref{sec:results:language-modeling-gpt2} show that Tversky feature sharing can result in a dramatic reduction of neural network parameter count while improving their performance.

\subsection{Data-Domain Visualization of Projection Layers}
\label{sec:method:data-domain-param-visualization}
We introduce a novel method of visualizing projection layer parameters such as the weights of fully connected layers, and the prototypes and features of Tversky projection layers to enable their interpretation in the same domain as data stimuli.
Our method is based on the hypothesis that those prototypes and features are concepts that should be recognizable in the data domain, and is applicable regardless of the depth at which those projection layers are employed in deep neural networks.
Our proposed visualization method consists of specifying the projection parameters as tensors of the same shape as the input data. These input-domain-specified prototypes and features are forwarded through the neural network just like data stimuli up to the layer prior to the layer in which they are used. Their obtained vector representation is subsequently used to perform the projection.
This approach significantly differs from prior approaches of interpreting neural network parameters and decisions based on 
the visualization of activations, 
optimization of input stimuli,
or construction of adversarial examples \citep{zeiler2014visualizing,mordvintsev2015inceptionism,selvaraju2016grad,zhang2018visual, zhang2020visualization,fan2021interpretability}.
% \todo{single layer visualization}
While our proposed method offers the clear advantage of visualizing projection parameters, it comes with the limitation that parameters specified in data-space are typically larger in size than their original counterparts, which increases the effective number of trainable parameters. These parameters also need to be forwarded along with every data batch, which induces additional training-time computation cost (See 
Figure~\ref{fig:data-domain-projection-param-visualization} in Appendix
\ref{sec:data-domain-projection-param-visualization}).
We use this technique to qualitatively compare the parameters of a linear projection layer and a Tversky projection layer of neural networks trained to recognize handwritten digits in our qualitative analysis experiment presented in Section~\ref{sec:results:qualitative-analysis}, which revealed that Tversky Projection layers's parameters are far more interpretable than the ones of the contemporary fullly connected layer (see Figure~\ref{fig:visual-prototypes-and-features}).

\section{Experiments}
\label{sec:quantitative-experiments}
Here we demonstrate the utility of our proposed differentiable Tversky similarity function in machine learning.
First, we show by construction that a single Tversky projection layer can model the {\sc xor} function and report empirical results on the same learning task (Section~\ref{sec:results:xor}).
Then, in Sections~\ref{sec:results:language-modeling-gpt2} and \ref{sec:results:image-recognition-resnet-50},  we conduct empirical experiments with state-of-the-art neural networks for language modeling and image recognition in which we compare baseline neural networks with their counterparts employing Tversky projection layers.
Finally, in Section~\ref{sec:results:qualitative-analysis}, we conduct qualitative analyses demonstrating the interpretability of Tversky neural networks.

% In this section, we also show that Tversky neural networks exhibit properties predicted by Tversky, including prototypicality and salience, and enable explaining similarity between objects in terms of their common and distinctive features.

% \ref{sec:results:xor}
% \ref{sec:results:image-recognition-resnet-50}
% \ref{sec:results:language-modeling-gpt2}
% \ref{sec:results:qualitative-analysis}

% Here we conduct experiments comparing state-of-the-art  image recognition and language modeling neural networks ResNet50 \citep{he2016deep} and GPT-2 \citep{radford2019language} to their counterparts employing Tversky projection layers.

\subsection{Modeling {\sc XOR} with a Single Tversky Projection Layer}
\label{sec:results:xor}
\begin{figure*}[ht]
    \centering
    \includegraphics[width=0.3\linewidth]{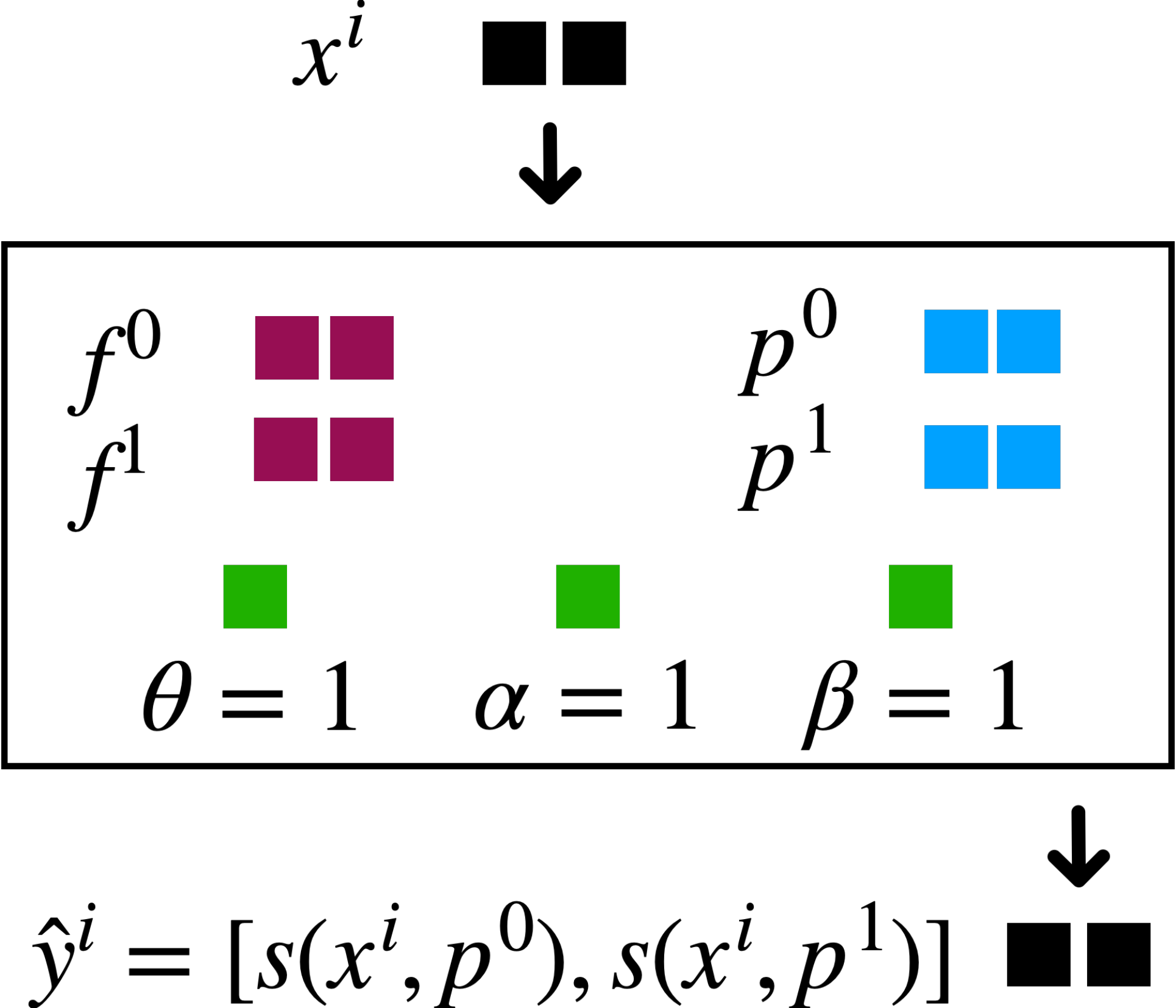}
    \quad
    \begin{tikzpicture}[scale=0.7]
        \definecolor{mpurple}{RGB}{151,14,83}
        \definecolor{mblue}{RGB}{0,161,255}

        % Draw axes
        \draw[->] (-2,0) -- (2,0) node[right] {};
        \draw[->] (0,-2) -- (0,2) node[above] {};
        
        % Draw grid
        \draw[very thin,color=gray] (-2,-2) grid (2,2);
    
        % Draw x vectors
        %\draw[->, very thick] (0,0) -- (0,0) node[below left] {$x^{(0)}$};
        \draw[->, very thick] (0,0) -- (0,1) node[left] {$x^1$};
        \draw[->, very thick] (0,0) -- (1,0) node[below] {$x^{2}$};
        \draw[->, very thick] (0,0) -- (1,1) node[above right] {$x^{3}$};
        
        % Draw vectors f1 and f2
        \draw[->, very thick, mpurple] (0,0) -- (0.5,-1) node[right] {$\mathbf{f^{0}}$};
        \draw[->, very thick, mpurple] (0,0) -- (-1,0.5) node[left] {$\mathbf{f^1}$};
        
        % Draw vectors p1 and p2
        \draw[->, very thick, mblue] (0,0) -- (-0.5,-0.5) node[below left] {$p^1$};
        \draw[->, very thick, mblue] (0,0) -- (0.5,0.5) node[above ] {$p^0$};
    
        % TODO compute actual orthogobal projection points
        \draw[-, dotted, red] (-0.5,-0.5) -- (-0.5,0.2);
        \draw[-, dotted, red] (-0.5,-0.5) -- (0.2,-0.2);
        \draw[-, dotted, red] (0,1) -- (-0.3,0.2);
        \draw[-, dotted, red] (1,0) -- (0.2,-0.3);
    \end{tikzpicture}
    \includegraphics[width=0.3\linewidth]{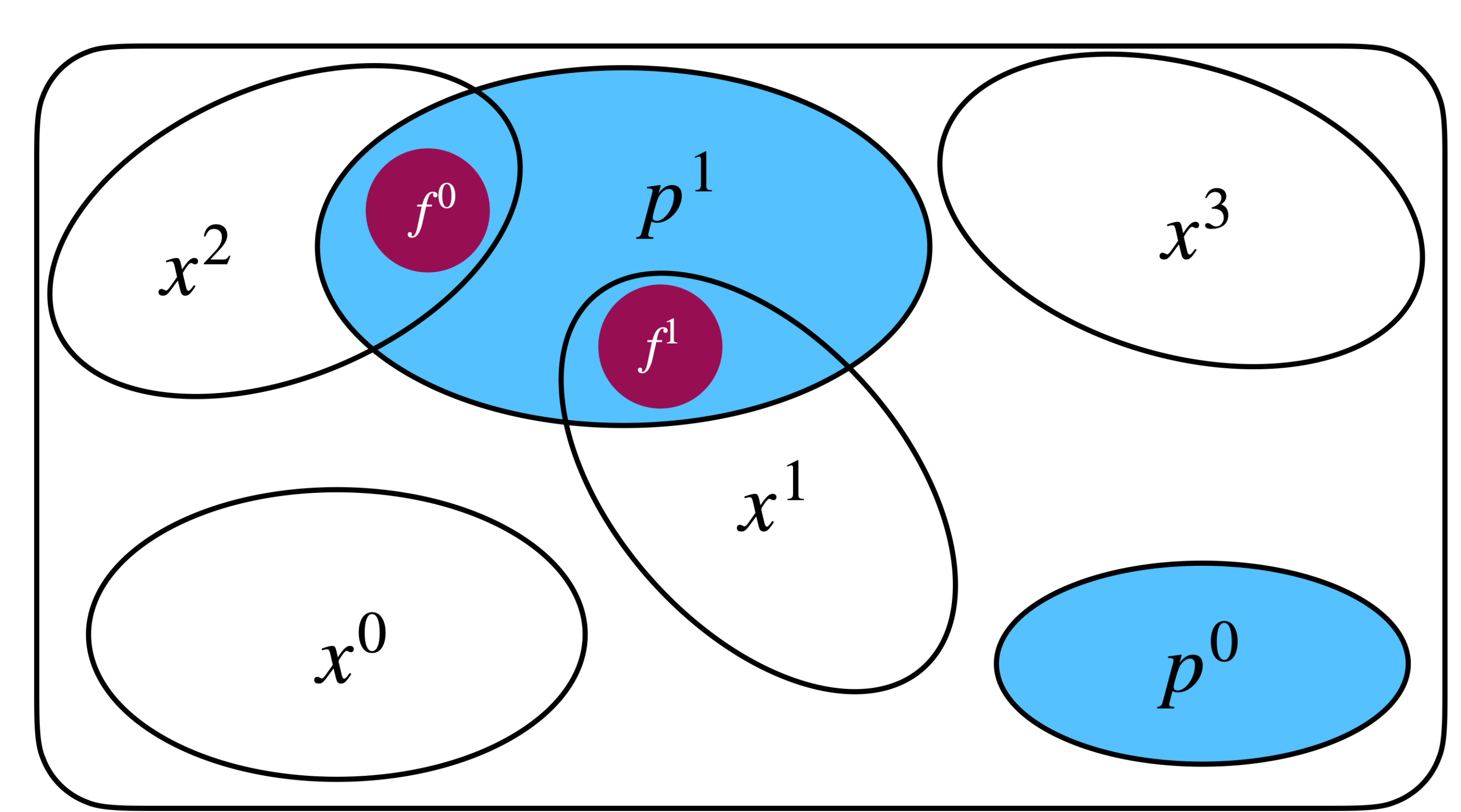}
    \caption{
    \textbf{(left)} A single tversky projection layer with 11 learnable parameters that computes the xor function. The input is a 2 digit binary number encoded as $x^i \in \mathcal{R}^2$. The output $\hat{y}^i$ is such that $txor(x^i) \iff s(x^i,p^1) > s(x^i,p^0)$, where $s$ is the proposed similarity function.
    \textbf{(middle)} Input vectors $x^i$ along with features $f^i$, and prototypes $p^i$ modeling the $xor$ function. The dot products
    $x^1 \cdot f^1$,
    $x^2 \cdot f^0$,
    $p^1 \cdot f^0$ and
    $p^1 \cdot f^1$ are positive.
    \textbf{(right)} Feature vectors define a universe in which input vectors
    $x^0=\{\}$, 
    $x^1=\{f^1\}$, 
    $x^2=\{f^0\}$, 
    $x^3=\{\}$,
    and learned prototype vectors
    $p^0=\{\}$, 
    $p^1=\{ f^0, f^1\}$, are represented as set of features. By convention, an object (input or prototype) contains a feature if its dot product with the feature vector is positive. By Tversky's contrast model of similarity,
     $s(x^3,p^0) > s(x^3,p^1)$, and $s(x^2,p^1) > s(x^2,p^0)$ therefore $txor([1, 1])=0$ and $txor([1, 0])=1$. Similarly for other inputs, $txor([0, 0])=0$ and $txor([0, 1])=1$.
    }
    \label{fig:constructed-tversy-xor}
\end{figure*}

As a first experiment, we construct a single Tversky projection layer that computes the {\sc xor} function, which is not computable by a single linear projection due to the required non-linear decision boundary.
Figure~\ref{fig:constructed-tversy-xor} shows the constructed projection, and its data and parameter vectors along with their set-centric interpretation.
In empirical experiments in which we train a Tversky projection to learn {\sc xor} with gradient descent under various hyperparameter conditions, we find that:
Some initializations of prototypes and features lead to convergence failure;
Initializing those parameters from a uniform distribution leads to higher convergence probability compared to normal and orthogonal initialization;
Normalizing prototype and object vectors deteriorates convergence probability;
\emph{product} and \emph{substractmatch} work best as values of the \emph{intersection reduction} and \emph{difference reduction} hyperparameters;
Convergence probability doesn't increase monotonically with the feature bank size;
and finally, Tversky projection can model {\sc xor} with as little as one feature. See Appendix~\ref{appendix:tversky-xornet}. 

\subsection{Language Modeling with TverskyGPT-2}
\label{sec:results:language-modeling-gpt2}
% \label{sec:language-modeling-experiment}
\begin{table}
    \centering
    \caption{Perplexity (PPL) and parameter counts (Params) of GPT2 and TverskyGPT2 trained for the language modeling task on the PTB dataset. Perplexities are reported on the held out test split.}
    
\begin{adjustbox}{max width=\textwidth}
    \begin{tabular}{rrrrrrrr}
        Model & Init & Tied Proto & Features  & Params & PPL & $\Delta Params$ & $\Delta PPL$  \\
        \toprule
            baseline 
                & finetuned 
                & N 
                &  
                & 163\,037\,184 
                & 30.52
                & 
                & 
                \\
            tversky-head 
                & finetuned 
                & N 
                & 16\,384   
                & 175\,620\,099 
                & \textbf{28.33}
                & +7.7\% % (175620099.0-163037184)/163037184
                & -7.2\% % (28.33-30.52)*100/30.52
                \\
        \midrule
            baseline 
                & finetuned 
                & Y 
                & 
                & 124\,439\,808  
                & \textbf{18.31}
                & 
                & 
                \\
            tversky-head 
                & finetuned 
                & Y 
                & 32\,768 
                & 149\,605\,635 
                & 18.62
                & +20.2\% %(149605635-124439808)*100/124439808
                & +1.7\%  %(18.62-18.31)*100/18.31
                \\
        \midrule
            baseline 
                & scratch 
                & N 
                & 
                & 163\,037\,184  
                & 111.79
                & 
                & 
                \\
            tversky-all-1layer
                & scratch 
                & N 
                & 4\,096 
                & 116\,591\,655 
                & \textbf{98.22}
                & -28.5 \% % (116591655-163037184)*100/163037184
                & -12.1 \% % (98.22-111.79)*100 / 111.79
                \\
        \midrule
            baseline 
                & scratch 
                & Y 
                & 
                & 124\,439\,808  
                & 112.81
                &
                &
                \\
            tversky-all-1layer
                & scratch 
                & Y 
                & 8\,192 
                & 81\,140\,007 
%                & 116.08 % prod
                & \textbf{103.99}
                & -34.8\% %(81140007-124439808)*100/124439808
                & -7.8 \% %(103.99 - 112.81)*100/112.81
                \\
        \bottomrule
    \end{tabular}
\end{adjustbox}
    \label{tab:ptb-language-modeling-results-summary}
\end{table}

% TPERPLEXITIES ON PTB TEST SET

%

\subsubsection{Task and Data}
We compare the baseline GPT-2 small \citep{radford2019language} model with its Tversky variants on the Penn Treebank (PTB) language modeling benchmark \citep{marcus-etal-1993-building}.
\subsubsection{Method}
The following TverskyGPT-2 variants were considered:
\emph{tversky-head}, which replaces GPT-2's language modeling head's stack of linear layers with a Tversky projection layer, and
\emph{tversky-all}, which also replaces GPT-2's intermediate attention blocks' output projections with Tversky projections.
Tversky feature sharing is employed across all Tversky projection layers.
For all models, we experiment with prototype tying (tied vs not tied) and initialization (random vs OpenAI's released weights) while training for 50 epochs on PTB's training set, and validating on PTB validation set. 
The baseline models, and best Tversky models are evaluated on PTB's held-out test set.

\subsubsection{Results}
Tversky language models matched or surpassed baseline perplexity in all settings except the one in which 
pre-trained weights and prototype tying are employed
(Table~\ref{tab:ptb-language-modeling-results-summary}). In that setting, Tversky prototypes and features are still randomly initialized, while the baseline fully connected layers are initialized from the pretrained token embedding matrix.
The perplexity gap between the best Tversky neural network and baseline was higher when training from scratch, with a 7.8\% reduction in perplexity when prototypes are tied to input embeddings, and a 12.1\% reduction when they are not. In the tied prototype setting, the best Tversky model also has 34.8\% fewer parameters.
See
Appendix~\ref{sec:gpt2-ptb-results} for validation results, and Table~\ref{tab:ptb-language-modeling-results-summary} for test results.
%%%%%%%%%%%%%%%%%%%%%%%

% \include{exp_006/tables_ptb_summary}
\iffalse

\paragraph{PTB}
Perplexity:
\begin{itemize}
    \item init=finetuned and tie-proto=N: (32.31 - 34.85)/34.85= -7.29 percent
    \begin{itemize}
        \item params: (175620099-163037184)/163037184 = 7.717 percent
    \end{itemize}

    \item init=finetuned and tie-proto=Y: (20.46 - 19.99)/19.99 = 2.35 percent
    \begin{itemize}
        \item params: (149605635-124439808)/124439808 = 20.22 percent
    \end{itemize}
    
    \item init=scratch and tie-proto=N.: (117.59 - 134.06)/134.06 = -12.29 percent
    \begin{itemize}
        \item params: (116591655 - 163037184)/163037184 = -28.49 percent
    \end{itemize}
    
    \item init=scratch and tie-proto=Y: (125.86-136.04 )/136.04 = -7.483 percent
    \begin{itemize}
        \item params: (81140007-124439808 )/124439808 = -34.79 
        \item see Table~\ref{tab:ptb-perplexity_init=scratch_tie_proto=Y}
    \end{itemize}
\end{itemize}

\fi

% Notes for experiment (100 not reported in current version)
% justification of NaBirds hparams: https://arxiv.org/pdf/1909.04412
%   - images resized 600 × 600
%   - image patches of size 448 × 448 from random cropping (training)
%   - augmentation flipped horizontally with aprobability of 0.5.
%   - center cropping 448 x 448 (testing)
%   -  For datasets that do not provide a validation set, we randomly take 10\% out of the training samples from each category for validation.
%   - We train the network for 30 epochs and decay the learning rate by 0.1 every 15 epochs
\subsection{Image Recognition with TverskyResNet50}
\label{sec:results:image-recognition-resnet-50}
\subsubsection{Task and Data}
We experiment with replacing the final linear projection layer in ResNet50 \citep{he2016deep} with a Tversky projection and report the accuracy of the baseline and Tversky variants on
MNIST \citep{lecun1998gradient} handwritten digits 
and NABirds \citep{van2015building} bird species classification tasks.

\subsubsection{Method}
ResNet-50 and TverskyResNet-50, which uses a Tversky Projection instead of the final fully connected layer, are trained with the same hyperparameters (200 epochs; batch sizes: (NABirds, 256), (MNIST, 1024); lr: 0.03; momentum 0.9; weight decay $10^{-8}$; dropout: 0.1; and gradient clipping threshold $10$). In the \emph{Pretrained backbone} condition, ResNet50 is initialized with pretrained ImageNet weights. In the \emph{Freeze Backbone} condition, the convolution layer weights are not updated during training. Tversky prototypes and features are randomly initialized in all settings.
Tversky feature intersection and difference reduction hyperparameters (\emph{product} and \emph{ignorematch}) were informed by our results in Section~\ref{sec:results:language-modeling-gpt2}. We chose Tversky feature bank sizes 20 and 224 for MNIST and NABirds respectively.
Unlike in experiment~\ref{sec:results:language-modeling-gpt2}, which employs a strict train/validation/test protocol, no extensive hyperparameter tuning, and validation was employed in this experiment because the employed datasets, despite having sizeable test sets, do not have official distinct validation sets.
Nonetheless the following results demonstrate the existence of settings under which Tversky vision models match or surpass baseline accuracy on official train/test splits.

\subsubsection{Results}
Classification accuracies reported in  table~\ref{tab:resnet-50-mnist-nabirds} show that Tversky vision models can match or surpass baseline accuracy under the specified experimental conditions. In particular, when Tversky projection was used as a domain adapter with a frozen backbone, $8.5\%$ and $24.7\%$ improvement over baseline was observed.
In this experiment, the Tversky neural networks have more parameters than the baseline, because their parameters also include a feature bank in addition to the prototype bank while the baseline only contains a prototype bank, corresponding to the weights of the fully connected layer. 
%Figure~\ref{fig:tversky-resnet-50-mnist} shows TverskyResNet-50's accuracy on MNIST in comparison to baseline with various feature counts. TverskyResNet-50's feature bank size is set to 224 in the NABirds experiments. 
%
While the increased parameter count of TverskyResNet-50 could contribute to the observed accuracy improvement, our results in Section~\ref{sec:results:language-modeling-gpt2} show that the use of Tversky neural networks can simultaneously result in decreased parameter count and increased performance compared to baseline.

%%%%%%%%%%%%%%%%%%%%%%%%%%%%%%%%%

% MNIST
% Improvement in all settings.
% Adaptation:  8.44 percent; 
% Finetuning: 0.16 percent;
% Scratch: 0.58 percent;
    
% Adaptation: (62.26 - 57.41) / 57.41 = 8.44 percent 
% Finetuning: (99.31-99.15)/99.15=0.16 percent
% Scratch: (98.60-98.03)/98.03 = 0.58 percent

% NABIRDS
% Table~\ref{tab:nabirds-results}.
% Improvemnet in all settings: Adaptation:  24.72 percent;
% Finetuning: 0.12 percent;
% Scratch: 7.18 percent;

% Adaptation: (44.9 - 36.0) / 36.0 = 24.72 percent 
% Finetuning: (83.7-83.6)/83.6=0.12 percent
% Scratch: (65.7 - 61.3)/61.3 = 7.18 percent

% external references for dnn accuracy on nabirds
% https://arxiv.org/pdf/2504.20322  76% to 84%
% https://openaccess.thecvf.com/content/ICCV2023W/NIVT/html/Haurum_Which_Tokens_to_Use_Investigating_Token_Reduction_in_Vision_Transformers_ICCVW_2023_paper.html 80%

\begin{table}
% note: MNIST: experiment 003
% note nabirds: experim,ent 001.02
    \centering
    \caption{Accuracy of Resnet-50 (Baseline) and Tversky-Resnet-50 (Tversky) on the tasks of MNIST handwritten digit classification and NABirds bird species classification. \emph{Pretrained} (True when weights are initialized from ImageNet, False when they are randomly initialized). \emph{Frozen} (True when only the final projection layer is finetuned, False when the entire model is finetuned)}
    \begin{tabular}{lccc|cc}
        \toprule
         &  & \multicolumn{2}{c}{MNIST} & \multicolumn{2}{c}{NABirds} \\
        \cmidrule(lr){3-4} \cmidrule(lr){5-6}
        Pretrained & Frozen & Tversky & Baseline  & Tversky & Baseline \\
        \midrule
        True & True & \textbf{62.3}  & 57.4  & \textbf{44.9}  & 36.0 \\
        True & False & \textbf{99.3} &  99.2 & \textbf{83.7} & 83.6 \\
%        False & True & 20.23 & \textbf{28.45}  \\
        False & False & \textbf{98.6} & 98.0 & \textbf{65.7} & 61.3  \\
        \bottomrule
    \end{tabular}
    
    \label{tab:resnet-50-mnist-nabirds}
\end{table}

\iffalse

\begin{table}
    \centering
    \begin{tabular}{lcccc}
        \toprule
        Pretrained & Freeze & \multicolumn{2}{c}{Accuracy (\%)} \\
        \cmidrule(lr){3-4}
        Backbone & Backbone & Tversky & Baseline \\
        \midrule
        True & True & \textbf{62.26}  & 57.41 \\
        True & False & \textbf{99.31} &  99.15 \\
%        False & True & 20.23 & \textbf{28.45}  \\
        False & False & \textbf{98.60} & 98.03  \\
        \bottomrule
    \end{tabular}
    \caption{Exp 003 (MNIST): Accuracy results for handwritten digit classification using the mnist dataset. The backbone is a Resnet-50. pretrained mean pretrained on imagenet. Freeze backbone means that the pretrained backbone is not updated during training.}
    \label{tab:resnet-50-mnist}
\end{table}

\begin{table}
    \centering
    \begin{tabular}{lcccc}
        \toprule
        Pretrained & Freeze & \multicolumn{2}{c}{Accuracy (\%)} \\
        \cmidrule(lr){3-4}
        Backbone & Backbone & Tversky & Baseline \\
        \midrule
        True & True & \textbf{44.9}  & 36.0 \\
        True & False & \textbf{83.7} & 83.6 \\
%        False & True & 00.9 & \textbf{02.2}  \\
        False & False & \textbf{65.7} & 61.3  \\
        \bottomrule
    \end{tabular}
    \caption{Exp 001.02: (Nabirds) Comparison of PsySim and Baseline model accuracies under different configurations}
    \label{tab:nabirds-results}
\end{table}

\fi

%%%%%%%%%%%%%%%%%%%%%%%%%%

\subsection{Qualitative Analysis of Tversky Neural Networks}
%\label{sec:quantitative-analysis}
\label{sec:results:qualitative-analysis}
% \label{sec:experiment:data-domain-param-visualization}

\begin{figure*}
    \centering
    \includegraphics[width=0.99\linewidth]{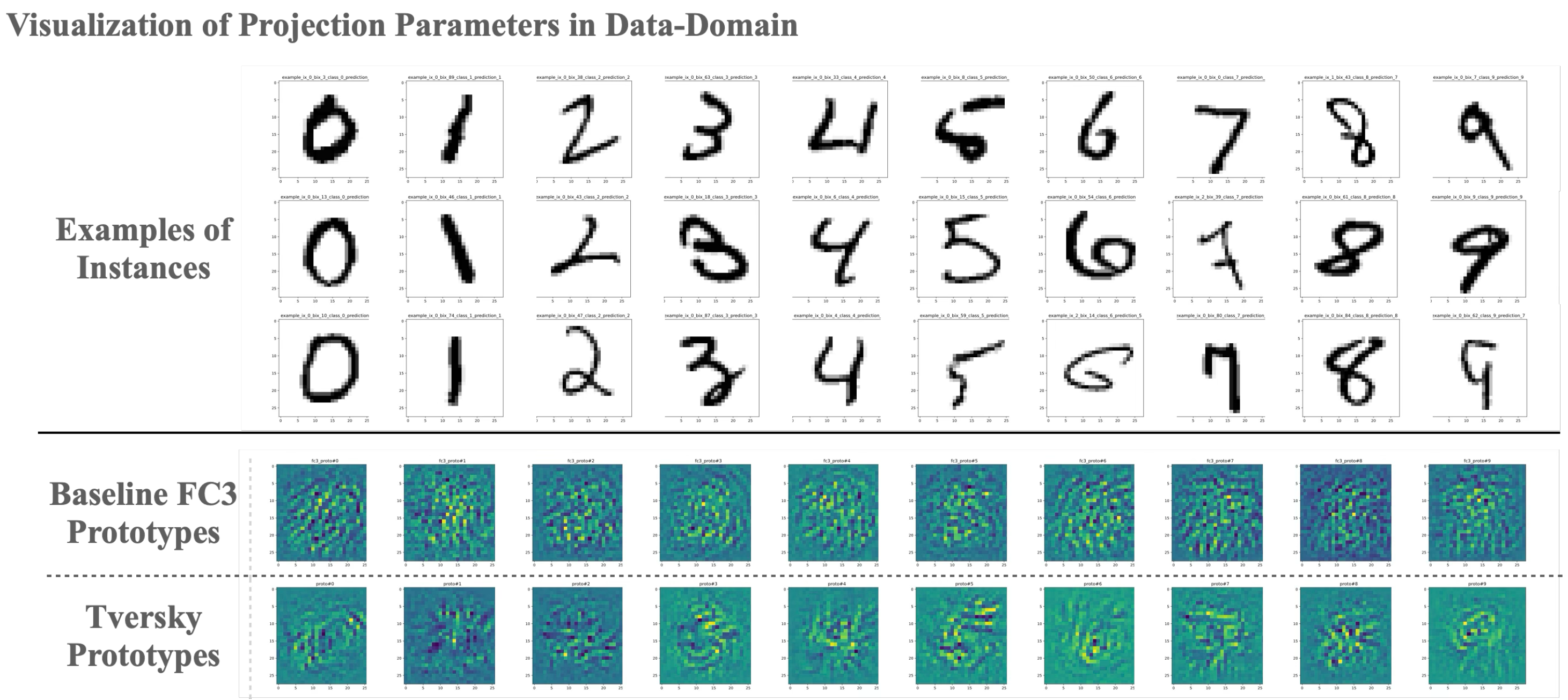}
    \caption{
        Visualization of prototypes using our input domain  projection parameter specification method.
        Top: 3 examples for each class. Bottom: 10 columns of the final linear projection layer of VisualMNISTNet, and 10 prototype vectors of the Tversky Projection Layer of TverskyMNISTNet.
        Both models achieve 99\%  accuracy on the validation set. Handwritten digit features are more perceptible to humans in Tversky prototypes and features compared to the linear projection prototypes.}
    \label{fig:visual-prototypes-and-features}
\end{figure*}

% \subsubsection{Task and Data}

Here we present results showing the interpretability of Tversky neural networks. 
We discuss the explainability of learned prototypes, the quantification of salience, and the algebraic specification of semantic fields enabled by Tversky neural networks.

\subsubsection{Method}
\textbf{Prototype Visualization:} 
To visually compare the prototypes learned by Tversky Neural Networks and classical neural networks, 
we employ the prototype visualization method described in Section~\ref{sec:method:data-domain-param-visualization} and illustrated in Appendix~\ref{sec:data-domain-projection-param-visualization}. 
This method is used to display the handwritten digit prototypes learned by VisualMNISTNet and VisualTverskyMNISTNet, 
two neural networks inspired by LeNet-5 \citep{lecun1998gradient}. 
Further details are provided in Appendix~\ref{sec:tversky-mnistnet}.

\textbf{Salience:}  
To assess whether salience, as computed using Equation~\ref{eq:saliency}, is intelligible to humans, 
we rank MNIST examples by their salience score.

\textbf{Semantic Fields:}  
Tversky Neural Networks' repesentation of objects as sets of features allows the algebraic specification of \emph{semantic fields}. 
A semantic field is a subset of features resulting from set expressions that are interpretable by humans. 
For instance, $A \cap B - C$ describes the semantic field capturing the common features of $A$ and $B$ that are distinct from $C$. 
To visualize a semantic field $F$, we rank examples $x_i$ by their score in that field:
$s = \sum_{f_k \in F}{f_k \cdot x_i}$.
This enables the visualization of semantics captured by TverskyMNISTNet's representation of images and TverskyGPT2 representation of tokens.

\subsubsection{Results}

\textbf{Prototypes:}  
Figure~\ref{fig:visual-prototypes-and-features} shows that Tversky prototypes exhibit stroke patterns similar to human handwriting, 
such as lines and curves, more clearly than those learned by linear projection layers, which exhibit texture patterns that are difficult to interpret. 
This lack of interpretability, even in this simple domain without background textures, represents a significant limitation of prior approaches.
Tversky prototypes are also more \emph{salient} than individual instances: 
they exhibit more features and appear to combine the features of all possible instances of their class. 
For example, the prototype for the handwritten ``7'' has both an upper-left serif and a middle horizontal stroke. 
The prototypes for the digit ``1'' and ``9'' include vertical lines slanted to the right and to the left. 
The prototype for class ``9'' combines both open and closed top loops.
Figure~\ref{fig:mnist-net-prototypes-over-iterations} shows the evolution of learned prototypes over training epochs for both VisualMNISTNet and TverskyVisualMNISTNet. 
We observed that $\alpha > \beta$ in trained Tversky neural networks (see Appendix~\ref{sec:example-of-learned-alpha-beta-theta}), 
indicating that they weigh the distinctive features of individual instances more heavily than those of prototypes, which is in line with Tversky's theory.

\begin{figure*}
    \begin{center}
    \includegraphics[width=0.9\textwidth]{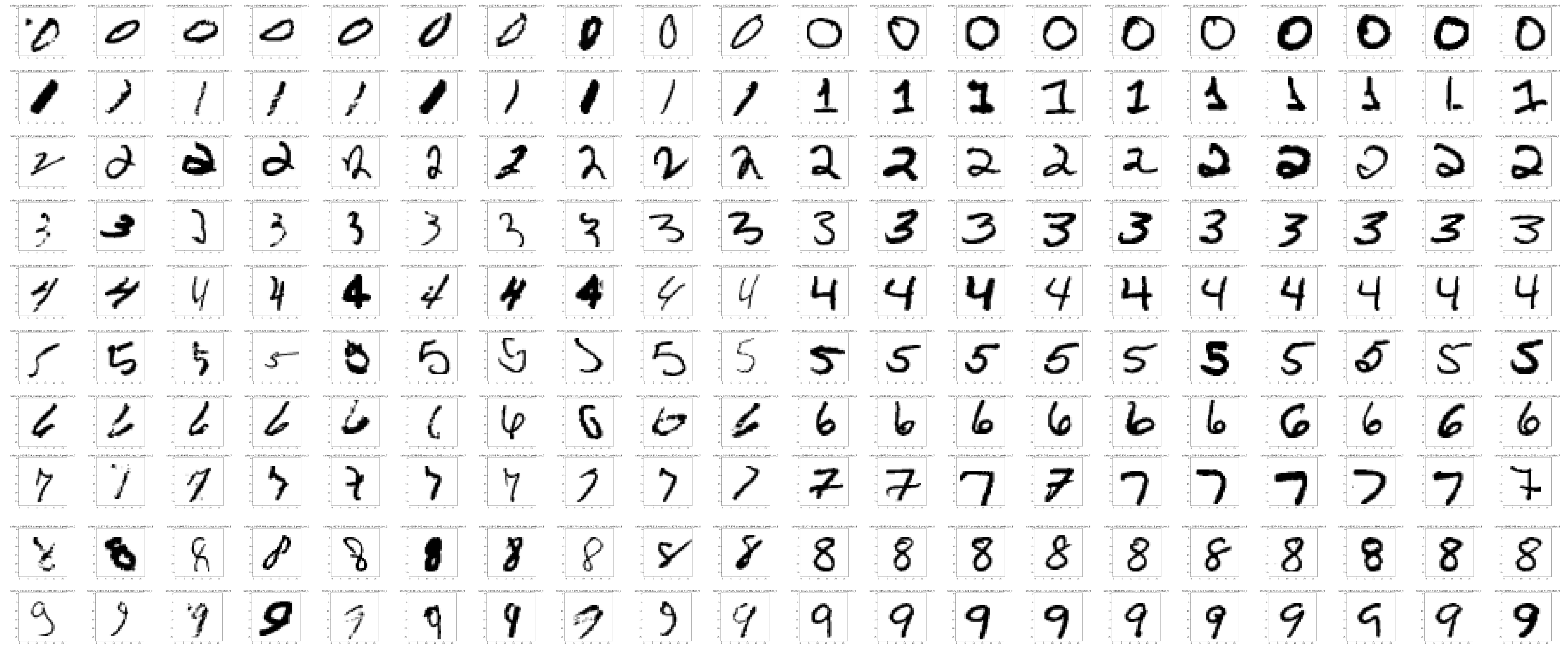}
    \end{center}
    \caption{MNIST digits sorted from low to high salience (Equation~\ref{eq:saliency}). Like humans, the underlying Tversky Neural Network perceives stimuli exhibiting more \emph{goodness of form} as more salient.}
    \label{fig:mnist-salience}
\end{figure*}

\textbf{Salience:}  
Figure~\ref{fig:mnist-salience} shows MNIST examples ranked from low to high salience according to Equation~\ref{eq:saliency}. 
This ranking aligns with Tversky's empirical psychological observations that humans perceive stimuli exhibiting more \emph{goodness of form} as more salient.

\textbf{Semantic Fields}
Table~\ref{tab:semantic-field-expression-examples} shows examples of semantic fields defined with TverskyGPT2 tokens, and top scoring tokens in those semantic fields. These results and additional results in Appendix~\ref{sec:gpt2-ptb-semantic-fields} show that interpretable semantic fields such as those capturing the concepts of \emph{adjectives}, \emph{comparatives}, \emph{superlatives}, \emph{verb forms}, or specific word \emph{senses}  can be specified algebraically with Tversky Language models.
Similar results can be achieved with Tversky vision models. Figure~\ref{fig:tversky-mnist-venn} shows two MNIST digits, and examples that illustrate their common and distinctive features.

% Compared to the parallelogram approach \citep{mikolov2013efficient,pennington2014glove,peterson2020parallelograms}, which is based on the geometric model of similarity.

\begin{table*}[ht] 
    \centering 
    
    \caption{Semantic concepts, their algebraic specification using set operations on TverskyGPT2 tokens represented as sets, and top tokens ranked by semantic score in the resulting semantic field. Tversky neural networks enable the algebraic specification of interpretable semantic concepts.} 

    \begin{tabularx}{\textwidth}{p{2.5cm}|p{4cm}|>{\raggedright\arraybackslash}X}
    \toprule
    Concept
    &
    Semantic Expression 
    & 
    Top-scoring tokens
    \\ \midrule 
    
    Adjective
    &
    $bad \cap good - worse - better - worst - best $
    & 
    \emph{lousy}, \emph{evil}, \emph{crappy}, \emph{poor}, \emph{shitty}, \emph{nice}, \emph{terrible}, \emph{horrible}, \emph{decent}, \emph{mediocre}, \emph{valid}, \emph{great}, \emph{excellent} 
    \\ \midrule 
    
    Comparative
    &
    $worse \cap better - bad - good -worst-best$
    &
    \emph{sharper}, \emph{nicer}, \emph{clearer}, \emph{smarter}, \emph{wiser}, \emph{smoother}, \emph{safer}, \emph{hotter}, \emph{happier}, \emph{tighter}, \emph{louder}, \emph{preferable} 
    \\ \midrule 
    
    Superlative
    &
    $worst \cap best - worse - better - bad - good$
    &
    \emph{happiest}, \emph{safest}, \emph{quickest}, \emph{deadliest}, \emph{finest}, \emph{hottest}, \emph{busiest}, \emph{toughest}, \emph{darkest}, \emph{fastest}, \emph{brightest}, \emph{holiest} 
    \\ \midrule 
    Industrial Plant:
    &
    $plant -aceae -vegetation -vegetable -planting -flower -herb -vine -crop -tree -mushroom -plantation -leaves $
    &
    \emph{facility}, \emph{stall}, \emph{brate}, \emph{factory}, \emph{Laboratories}, \emph{Shed}, \emph{implant}, \emph{Berm}, \emph{Sew}, \emph{Manufacturing}, \emph{reactor}, \emph{Diesel}, \emph{refinery}, \emph{strain}, \emph{distribut}, \emph{planting}, \emph{Industrial}, \emph{microbial}, \emph{actory}, \emph{reactors}, \emph{Unit}, \emph{Install}, \emph{depot}, \emph{spill}, \emph{Cutter}, \emph{Indust}, \emph{Bot}, \emph{Nuclear}, \emph{ineries} 
    \\ \bottomrule
    \end{tabularx} 
    
    \label{tab:semantic-field-expression-examples} 
\end{table*}

%
%Tversky prototypes enable humans to interpret the concepts learned by the neural network.

\section{Discussion}

\paragraph{Hyperparameter Tuning:}  
Our language modeling experiments suggest the existence of an optimal feature bank size, though determining this value \emph{a priori} remains an open question. 
In the XOR setting, we observed that Tversky networks can be sensitive to hyperparameters, underscoring the importance of standard tuning practices. 
For our vision experiments, no tuning was performed; our goal was simply to demonstrate the existence of Tversky models that can match or surpass baseline accuracy. 
We encourage practitioners to apply standard hyperparameter tuning to further improve performance, especially since Tversky networks are a novel architecture, whereas the baselines we compare against have benefited from years of community-driven refinement.

\iffalse
\textbf{Tversky neural networks can be more parameter-efficient than baseline.}
When trained from scratch while tying language modeling head prototypes, 
Tversky-GPT2 employing Tversky projections after intermediate attention blocks and in the language modeling head achieved 7.5\% fewer perplexity with 34.8\% fewer parameters (Table~\ref{tab:ptb-language-modeling-results-summary}).
%
TverskyMNISTNet achieves similar accuracy as MNISTNet with 3 times less parameters (Table~\ref{tab:mnist-interpretability-experiments}). 
\fi

\textbf{Weight tying in language models is adequate.}
Our experiments with GPT-2 on PTB showed that weight tying can be beneficial in both language modeling heads employing Tversky and linear projections.
Our interpretation of projection layers as measurement of similarity of stimuli to prototypes explains the adequacy of weight tying in language models: Token embeddings are prototypes, and the language modeling head classifies output tokens by measuring their similarity to those prototypes. This practice with Tversky projection layers is further principled given the underlying theory of similarity.

\textbf{Data-domain projection parameter specification could enable interpretable editing.}
For instance, excluding handwritten sevens with the mid horizontal stroke from VisualTverskyMNISTNet's training set would result in a prototype without that stroke, and high error rates on handwritten sevens with that stroke. The resulting model could be edited by adding a feature specifying the stroke, and editing the prototype to include the stroke. This neural network editing approach could enable the understanding and eradication of dangerous model biases.

\section*{Conclusion}
This work introduces a differentiable similarity function based on a psychologically plausible theory of similarity.
The Tversky similarity layer and Tversky projection layer serve as basic neural network building blocks implementing this theory.
We illustrated the Tversky projection layer in simplified settings that permit qualitative analysis, as well as in larger-scale image recognition and language modeling tasks.
Our experiments provide compelling evidence that our introduced neural modules are suitable for psychologically plausible deep learning while offering principled explainability, performance gains and parameter efficiency.

\iffalse
 Acknowledgements should only appear in the accepted version.

\section*{Acknowledgements}
TODO

\section*{Impact Statement}
TODO
Authors are \textbf{required} to include a statement of the potential 
broader impact of their work, including its ethical aspects and future 
societal consequences. This statement should be in an unnumbered 
section at the end of the paper (co-located with Acknowledgements -- 
the two may appear in either order, but both must be before References), 
and does not count toward the paper page limit. In many cases, where 
the ethical impacts and expected societal implications are those that 
are well established when advancing the field of Machine Learning, 
substantial discussion is not required, and a simple statement such 
as the following will suffice:

``This paper presents work whose goal is to advance the field of 
Machine Learning. There are many potential societal consequences 
of our work, none which we feel must be specifically highlighted here.''

The above statement can be used verbatim in such cases, but we 
encourage authors to think about whether there is content which does 
warrant further discussion, as this statement will be apparent if the 
paper is later flagged for ethics review.
\fi


\clearpage
\bibliographystyle{unsrtnat}
\bibliography{main}

%%%%%%%%%%%%%%%%%%%%%%%%%%%%%%%%%%%%%%%%%%%%%%%%%%%%%%%%%%%%%%%%%%%%%%%%%%%%%%%
%%%%%%%%%%%%%%%%%%%%%%%%%%%%%%%%%%%%%%%%%%%%%%%%%%%%%%%%%%%%%%%%%%%%%%%%%%%%%%%
% APPENDIX
%%%%%%%%%%%%%%%%%%%%%%%%%%%%%%%%%%%%%%%%%%%%%%%%%%%%%%%%%%%%%%%%%%%%%%%%%%%%%%%
%%%%%%%%%%%%%%%%%%%%%%%%%%%%%%%%%%%%%%%%%%%%%%%%%%%%%%%%%%%%%%%%%%%%%%%%%%%%%%%
\newpage
\appendix
\onecolumn

\clearpage
\newpage
\section{Modeling 2 Digits Binary Addition with a Tversky Projection Layer}

\begin{figure*}[h]
    \centering
    \includegraphics[width=0.2\linewidth]{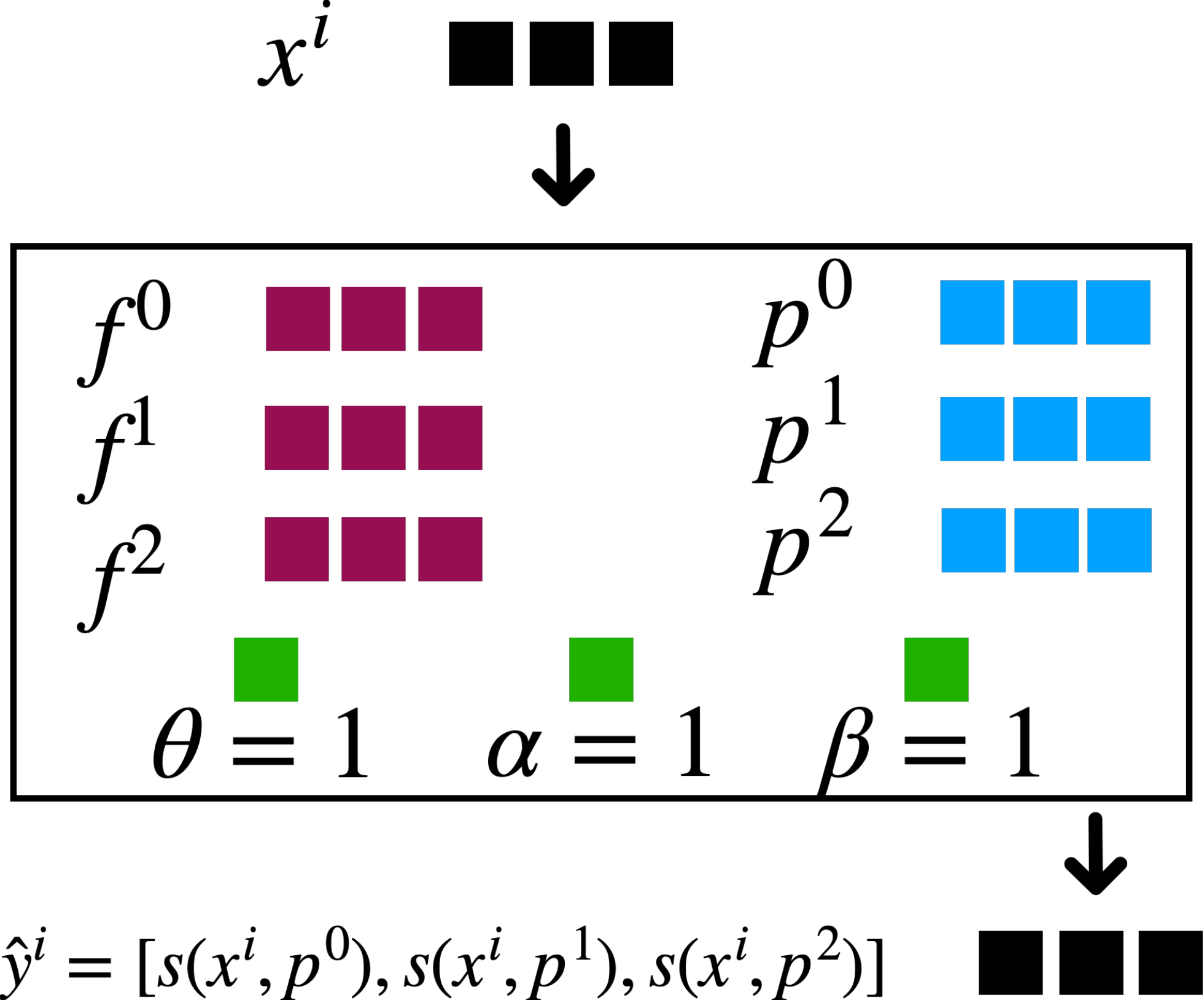}
        \begin{tikzpicture}[scale=1,>=latex]
        \definecolor{mpurple}{RGB}{151,14,83}
        \definecolor{mblue}{RGB}{0,161,255}
        
        % Define 3D coordinate system
        \begin{scope}[x={(1cm,0cm)}, y={(0.3cm,0.5cm)}, z={(0cm,1cm)}]
            
            % Draw coordinate axes
            \draw[->] (-2,0,0) -- (2,0,0) node[right] {x};
            \draw[->] (0,-2,0) -- (0,2,0) node[right] {y};
            \draw[->] (0,0,-2) -- (0,0,2) node[above] {z};
            
            % Draw grid on xy-plane (z=0)
            \foreach \x in {-2,-1,0,1,2}
                \foreach \y in {-2,-1,0,1,2}
                    \draw[very thin,gray] (\x,\y,0) -- (\x,{\y+1},0) -- ({\x+1},{\y+1},0);
            
            % Draw original x vectors (in xy-plane, z=0)
            \draw[->, very thick] (0,0,0) -- (0,1,0) node[left] {$x^1$};
            \draw[->, very thick] (0,0,0) -- (1,0,0) node[below] {$x^2$};
            \draw[->, very thick] (0,0,0) -- (1,1,0.5) node[above right] {$x^3$};
            
            % Draw original f vectors (in xy-plane, z=0)
            \draw[->, very thick, mpurple] (0,0,0) -- (0.5,-1,0) node[right] {$\mathbf{f^0}$};
            \draw[->, very thick, mpurple] (0,0,0) -- (-1,0.5,0) node[left] {$\mathbf{f^1}$};
            
            % Draw new f2 vector (0,0,1)
            \draw[->, very thick, mpurple] (0,0,0) -- (0,0,1) node[above right] {$\mathbf{f^2}$};
            
            % Draw original p vectors (in xy-plane, z=0)
            \draw[->, very thick, mblue] (0,0,0) -- (-0.5,-0.5,0) node[below left] {$p^1$};
            \draw[->, very thick, mblue] (0,0,0) -- (0.5,0.5,0) node[above] {$p^0$};
            \draw[->, very thick, mblue] (0,0,0) -- (1,1,1) node[above] {$p^2$};
            
            % Dotted projection lines
            \draw[-, dotted, red] (-0.5,-0.5,0) -- (-0.5,0.2,0);
            \draw[-, dotted, red] (-0.5,-0.5,0) -- (0.2,-0.2,0);
            \draw[-, dotted, red] (0,1,0) -- (-0.3,0.2,0);
            \draw[-, dotted, red] (1,0,0) -- (0.2,-0.3,0);
        \end{scope}
    \end{tikzpicture}
    \includegraphics[width=0.3\linewidth]{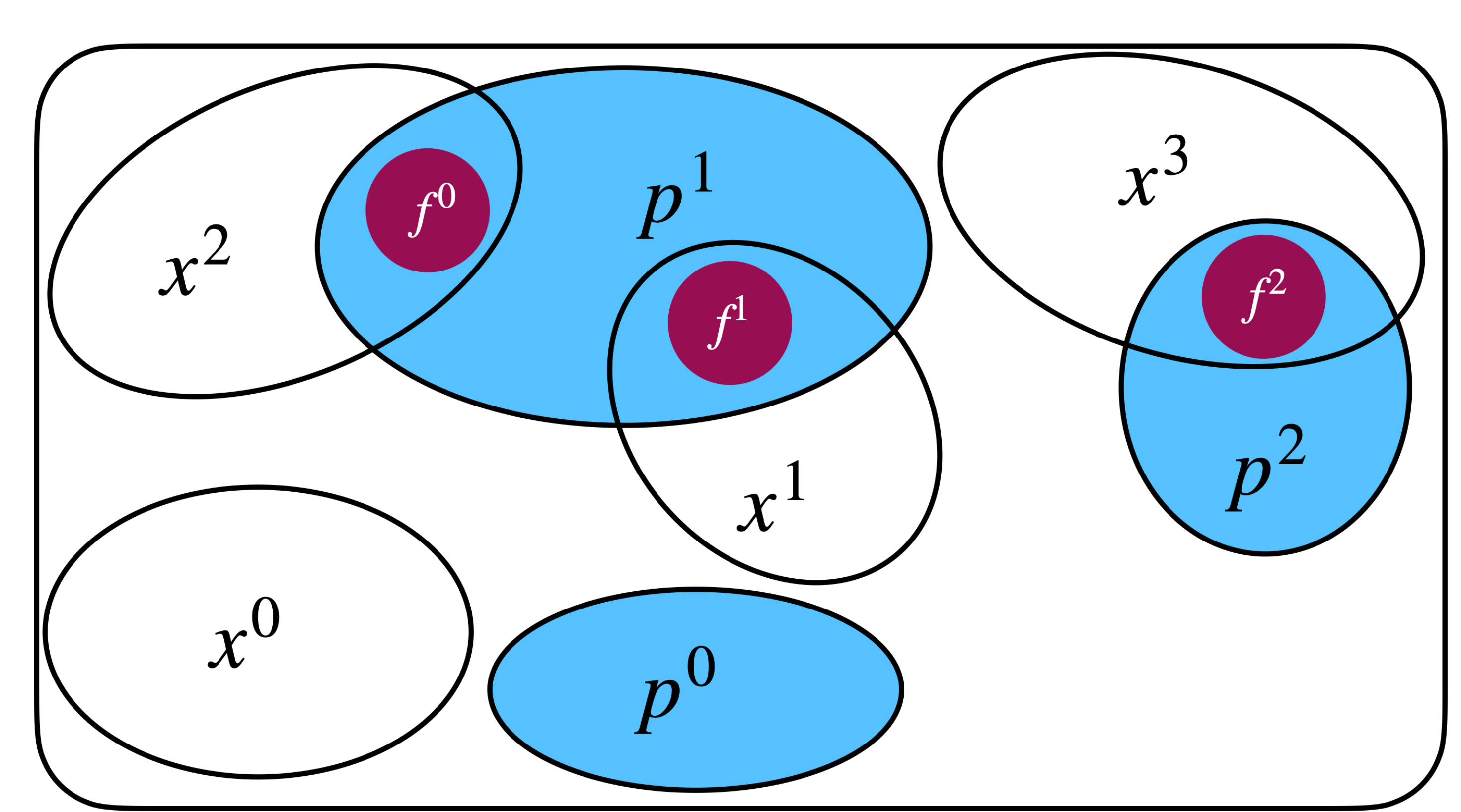}
    \caption{A tversky layer that adds 2 binary digits. Inputs are encoded in $\mathcal{R}^3$ as $x^0=[0,0,0]$, $x^1=[0,1,0]$, $x^2=[1,0,0]$ and $x^3=[1,1,1]$. The result of the addition corresponds to the prototype to which the input is most similar.   Compared to the txor model, this model employs one additional dimention to enable the introduction of a  feature $f^2=[0,0,1]$ only shared by $x^3$ and $p^2=[1,1,0.5]$. In this configuration, we have $tadd([0, 0])=0$, $tadd([0, 1])=tadd([1,0])=1$ and $tadd([1,1])=2$}
    \label{fig:enter-label}
\end{figure*}

\clearpage
\newpage
\section{Illustration of Data-Domain Visualization of Projection Layers}
\label{sec:data-domain-projection-param-visualization}

\begin{figure}[h]
    \centering
    \includegraphics[width=0.9\linewidth]{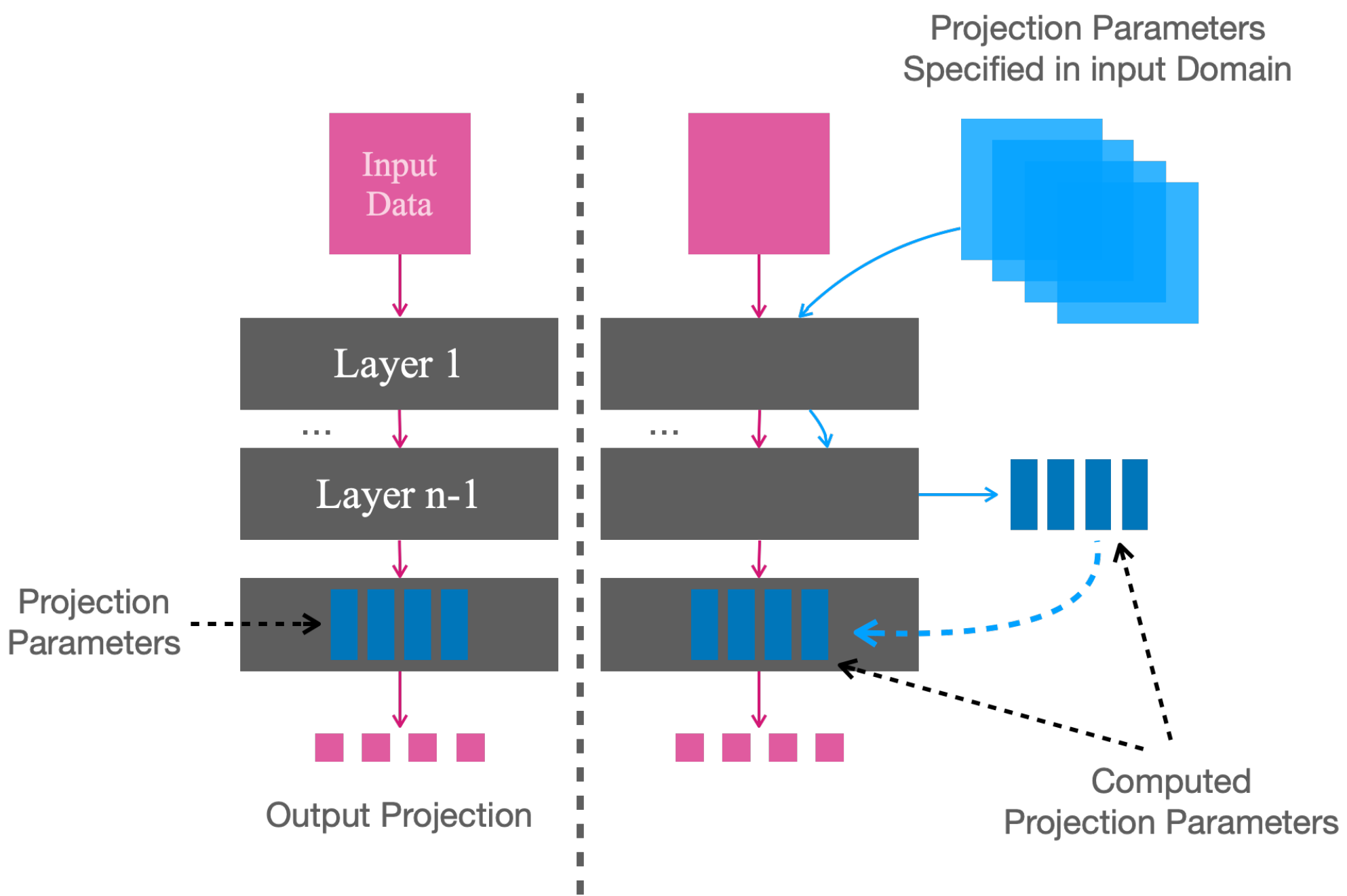}
    \caption{Illustration of our proposed data-domain visualization of projection parameters. \textbf{Left:} Classical deep neural network with a deep projection layer (Layer n) and its projection parameters illustrated as blue vectors. \textbf{Right:} Our proposed method. Projection parameters are specified as tensors of the same shape as the input data, and forwarded through the neural network up to layer n-1. The obtained vectors are used as projection parameters in Layer n. Using this method, the effective neural network parameter count is larger. However, this method enables visualizing the projection parameters in the input domain.}
    \label{fig:data-domain-projection-param-visualization}
\end{figure}

\clearpage
\newpage
\section{Training a Tversky Projection to Learn {\sc XOR} with Gradient Descent}
\label{appendix:tversky-xornet}

\subsection{Task and Data}
\begin{figure*}[ht]
    \centering
    \includegraphics[width=0.9\linewidth]{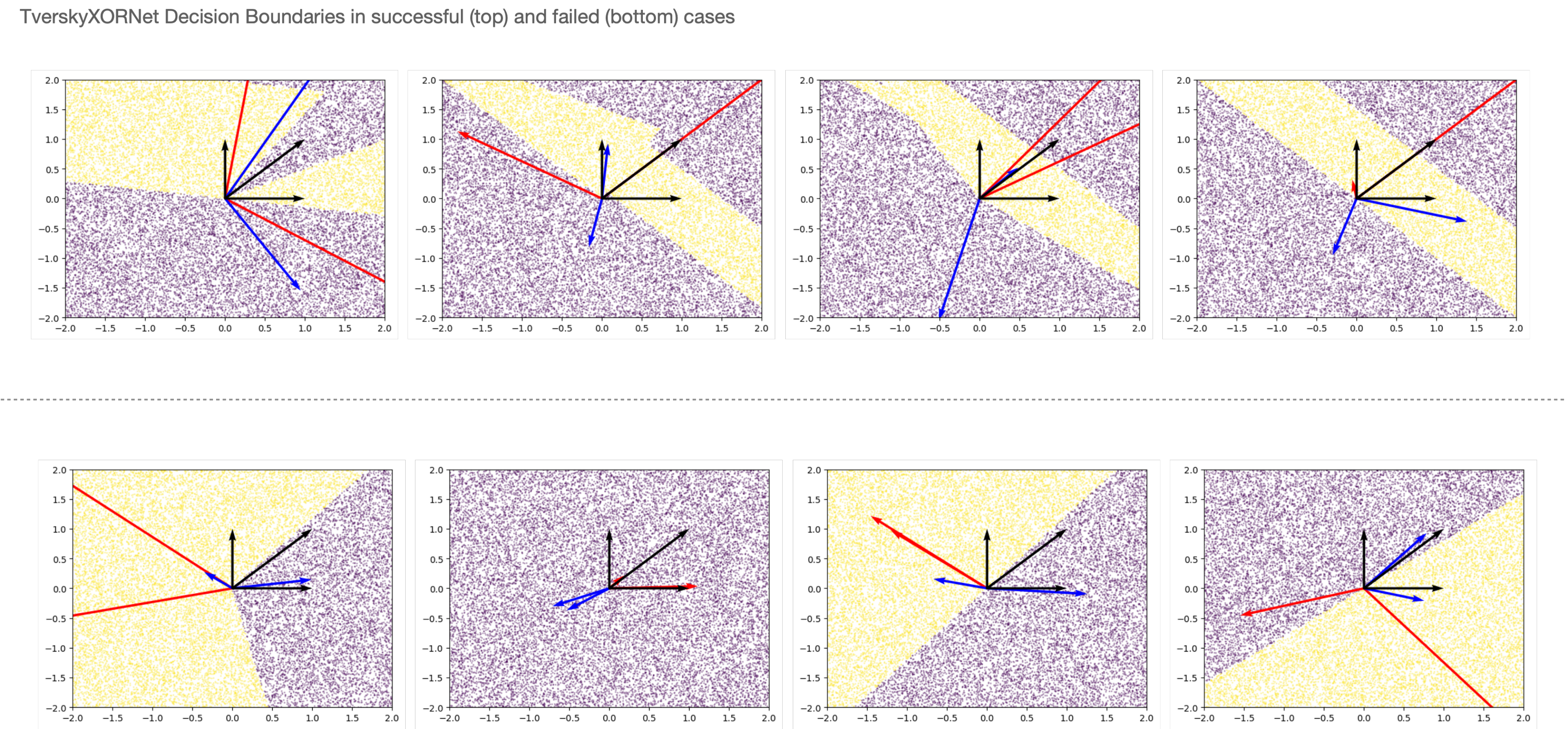}

    \caption{Example of learned TverskyProjection layers optimized to model the xor function. While multiple solutions are possible (top row), some initializations do not lead to convergence (botton row). The decision boundaries were drawn as follows.
Random input vectors [x,y] are uniformly sampled in the range [-2,-2] to [2,2] the tip of each vector is ploted as a colored dot, with the color representing the trained Tversky Projection layer's output. Tversky projection layers modeling the $xor$ function, which has a boolean domain and range extend $xor$ to the real vector space. Successful models should show data points [0,0] and [1,1] in purple (class 0), and [0,1] and [1,0] in yellow (class 1)}
    \label{fig:tversky-xornet-convergence}
\end{figure*}

In this experiment, we train a single Tversky projection layer to learn the {\sc xor} function via gradient descent under various hyper-parameter conditions.
Figure~\ref{fig:tversky-xornet-convergence} 
% in Appendix~\ref{appendix:tversky-xornet} 
shows that some initializations of Tversky projection layers may lead to gradient descent optimization failure.
%similarly to the baseline linear projection layer
This experiment empirically analyses its sensitivity to hyperparameters when trained to model the {\sc xor} function.

\subsection{Method}
To empirically estimate the sensitivity of Tversky projection layer's convergence to its hyperparameters, we train $12\,960$ xor models consisting of the following combinations of hyperparameters:
\begin{itemize}
    \item 6 intersection reduction methods \{\emph{min}, \emph{max}, \emph{product}, \emph{mean}, \emph{gmean}, \emph{softmin}\}; 
    \item 2 difference reduction methods \{\emph{ignorematch}, \emph{substractmatch}\};
    \item 2 normalization modes: \{false, true\};
    \item 6 feature counts: \{1, 2, 4, 8, 16, 32\};
    \item 3 prototype initialization distributions \{uniform, normal, orthogonal\};
    \item 3 feature bank initialization distributions \{uniform, normal, and orthogonal \};
    \item 9 random seeds.
\end{itemize}
Each model is trained for 1000 epochs. Tables \ref{tab:xor-inter-diff-aggregates}, \ref{tab:xor-init-aggregates}, \ref{tab:xor-normalize-aggregates}, and \ref{tab:xor-fbank-size-aggregates} reports the average and standard error of the convergence indicator (whether accuracy is 100\%) of the trained models marginalized by various combinations of hyperparameters. We refer to this variable as convergence probability $p(conv)$ in our results.

\subsection{Results}

\paragraph{Initialization of prototypes and features}
Initializing both features and prototypes by sampling from the uniform distribution resulted in the highest convergence probability (Table \ref{tab:xor-init-aggregates}).

\paragraph{Reduction of measures of intersections and differences}
Using \emph{product} and \emph{substractmatch} resulted in the highest convergence probability. (Table \ref{tab:xor-inter-diff-aggregates})

\paragraph{Normalization}
In this experiment, normalizing prototype and object vectors prior to calculating Tversky similarity decreased the convergence probability. (Table \ref{tab:xor-normalize-aggregates})

\paragraph{Feature count}
Tversky projection layer successfully modeled $xor$ with as little as 1 feature. $p(conv)$ was maximal with 16 features, but not monotonously increase with feature count.

\begin{figure*}[ht]
    \centering

    \includegraphics[width=0.9\linewidth]{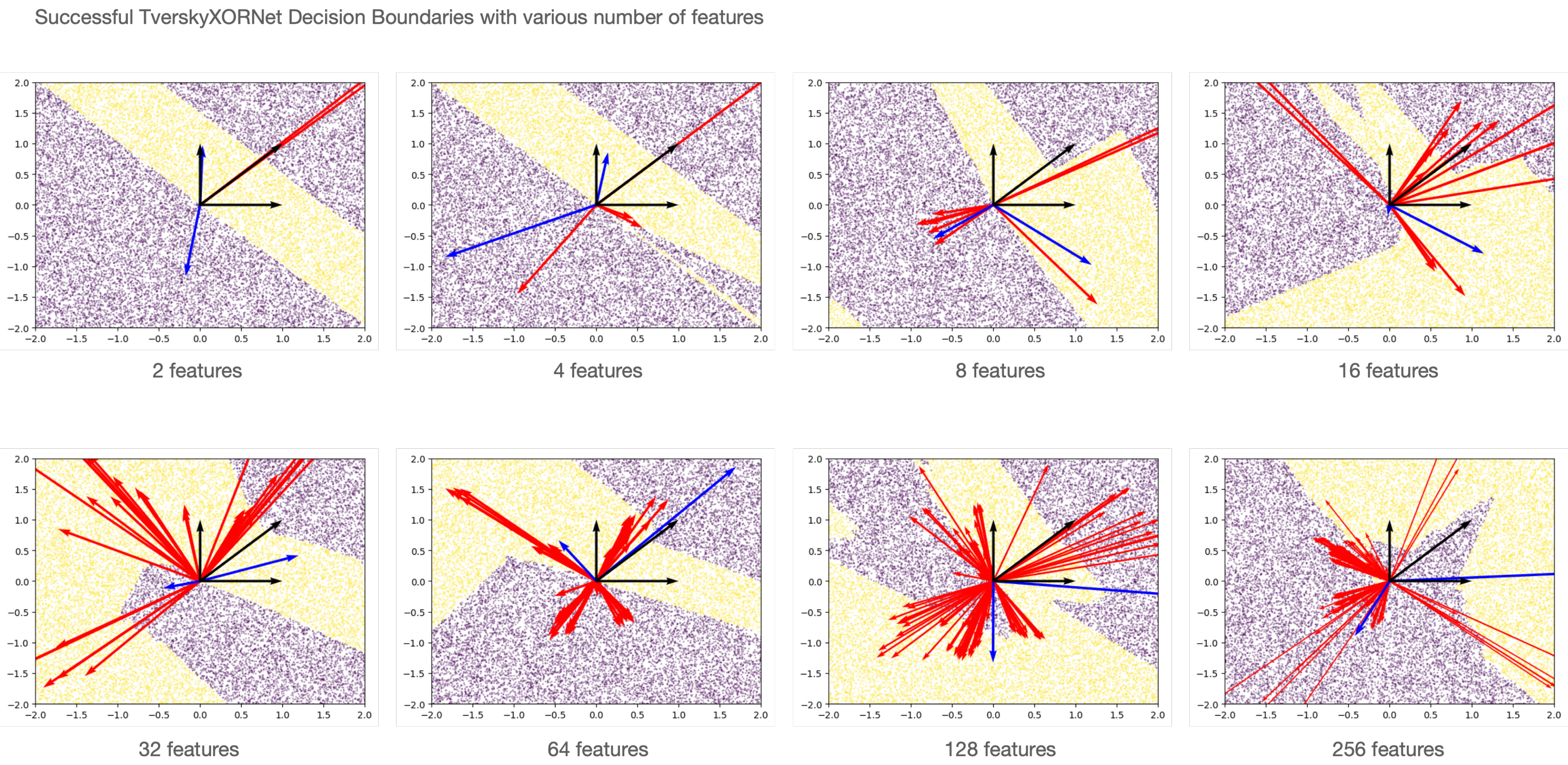}

    \caption{Example of successful Tversky decision boundaries modeling the XOR function with 2, 4, 8, 16, 32, 64, 128 and 256 features. The presence of clusters of features are frequent in overparameterized models.}
    \label{fig:xornet-with-various-features}
\end{figure*}

While linear projection layers cannot learn non-linear decision boundaries without composition with non-linear \emph{activation} functions, Figure~\ref{fig:xornet-with-various-features} shows that a single Tversky projection layers can learn complex non-linear decision boundaries even in low-dimensional vector space.

\begin{table}[ht]
\centering
\caption{Mean $\pm$ std error of loss and accuracy for TverskyXORNet with 
6 intersection reduction methods ($A \cap B$) and 
2 difference reduction methods ($A - B$).
Each row corresponds to the aggregate of 3x3x2x6x9 = 972 training sessions with:
3 fbank and prototype initialization methods(orthogonal, normal, uniform), 
2 normalization modes (normalized, and not normalized), 
6 feature bank sizes (1, 2, 4, 8, 16, 32), and
9 random seeds.
In each case, models were trained for 1000 epochs
}
\begin{tabular}{llrrrrr}
\toprule
$A \cap B$ & $A - B$ & n & loss & acc & best acc & p(conv)\\
\midrule
product & substractmatch & 972 & $0.27 \pm 0.01$ & $0.83 \pm 0.01$ & 1.0 & $0.53 \pm 0.02$\\
mean & substractmatch & 972 & $0.28 \pm 0.01$ & $0.82 \pm 0.01$ & 1.0 & $0.51 \pm 0.02$\\
max & ignorematch & 972 & $0.35 \pm 0.01$ & $0.78 \pm 0.01$ & 1.0 & $0.47 \pm 0.02$\\
max & substractmatch & 972 & $0.32 \pm 0.01$ & $0.80 \pm 0.01$ & 1.0 & $0.44 \pm 0.02$\\
softmin & substractmatch & 972 & $0.35 \pm 0.01$ & $0.80 \pm 0.01$ & 1.0 & $0.42 \pm 0.02$\\
min & ignorematch & 972 & $0.31 \pm 0.01$ & $0.78 \pm 0.01$ & 1.0 & $0.42 \pm 0.02$\\
softmin & ignorematch & 972 & $0.34 \pm 0.01$ & $0.78 \pm 0.01$ & 1.0 & $0.38 \pm 0.02$\\
mean & ignorematch & 972 & $0.42 \pm 0.01$ & $0.73 \pm 0.01$ & 1.0 & $0.26 \pm 0.01$\\
product & ignorematch & 972 & $0.41 \pm 0.01$ & $0.75 \pm 0.01$ & 1.0 & $0.23 \pm 0.01$\\
min & substractmatch & 972 & $0.50 \pm 0.00$ & $0.68 \pm 0.00$ & 1.0 & $0.02 \pm 0.00$\\
gmean & ignorematch & 972 & $nan \pm nan$ & $0.50 \pm 0.00$ & 0.5 & $0.00 \pm 0.00$\\
gmean & substractmatch & 972 & $nan \pm nan$ & $0.50 \pm 0.00$ & 0.5 & $0.00 \pm 0.00$\\
\bottomrule
\end{tabular}

\label{tab:xor-inter-diff-aggregates}
\end{table}

\begin{table}[ht]
\centering
\caption{
Mean $\pm$ std error of loss and accuracy for TverskyXORNet with 
3 fbank and prototype initialization methods (\emph{fbank init}, and \emph{proto init}).
Each row corresponds to the aggreate of 6x2x2x6x9 = 1,296 training sessions with:
6 intersection reduction methods (min, max, product, mean, gmean, softmin) and 
2 difference reduction methods (ignorematch, substractmatch),
2 normalization modes (normalized, and not normalized), 
6 feature bank sizes (1, 2, 4, 8, 16, 32), and
9 random seeds.
In each case, models were trained for 1000 epochs.
}
\begin{tabular}{llrrrrr}
\toprule
fbank init & proto init & n & loss & acc & best acc & p(conv)\\
\midrule
uniform & uniform & 1296 & $0.27 \pm 0.01$ & $0.78 \pm 0.01$ & 1.0 & $0.41 \pm 0.01$\\
uniform & normal & 1296 & $0.36 \pm 0.01$ & $0.72 \pm 0.01$ & 1.0 & $0.34 \pm 0.01$\\
normal & uniform & 1296 & $0.34 \pm 0.01$ & $0.74 \pm 0.01$ & 1.0 & $0.32 \pm 0.01$\\
normal & normal & 1296 & $0.35 \pm 0.01$ & $0.74 \pm 0.01$ & 1.0 & $0.31 \pm 0.01$\\
uniform & orthogonal & 1296 & $0.39 \pm 0.01$ & $0.70 \pm 0.01$ & 1.0 & $0.30 \pm 0.01$\\
orthogonal & uniform & 1296 & $0.37 \pm 0.01$ & $0.72 \pm 0.01$ & 1.0 & $0.29 \pm 0.01$\\
normal & orthogonal & 1296 & $0.37 \pm 0.01$ & $0.73 \pm 0.01$ & 1.0 & $0.28 \pm 0.01$\\
orthogonal & normal & 1296 & $0.37 \pm 0.01$ & $0.73 \pm 0.01$ & 1.0 & $0.26 \pm 0.01$\\
orthogonal & orthogonal & 1296 & $0.39 \pm 0.01$ & $0.71 \pm 0.01$ & 1.0 & $0.24 \pm 0.01$\\
\bottomrule

\end{tabular}

\label{tab:xor-init-aggregates}
\end{table}

\begin{table}[ht]
\centering
\caption{
Mean $\pm$ std error of loss and accuracy for TverskyXORNet with 
2 normalization modes (normalized, and not normalized).
Each row corresponds to the aggreate of 3x3x6x2x6x9 = 5832 training sessions with:
3 fbank and prototype initialization methods(orthogonal, normal, uniform).
6 intersection reduction methods (min, max, product, mean, gmean, softmin) and 
2 difference reduction methods (ignorematch, substractmatch),
6 feature bank sizes (1, 2, 4, 8, 16, 32), and
9 random seeds.
In each case, models were trained for 1000 epochs.
}
\begin{tabular}{lrrrrr}
\toprule
normalize & n & loss & acc & best acc & p(conv)\\
\midrule
False & 5832 & $0.33 \pm 0.00$ & $0.74 \pm 0.00$ & 1.0 & $0.34 \pm 0.01$\\
True & 5832 & $0.38 \pm 0.00$ & $0.72 \pm 0.00$ & 1.0 & $0.27 \pm 0.01$\\
\bottomrule
\end{tabular}
\label{tab:xor-normalize-aggregates}
\end{table}

\begin{table}[ht]
\centering
\caption{
Mean $\pm$ std error of loss and accuracy for TverskyXORNet with 
6 feature bank sizes (1, 2, 4, 8, 16, 32).
Each row corresponds to the aggreate of 3x3x6x2x2x9 = 1944 training sessions with:
3 fbank and prototype initialization methods(orthogonal, normal, uniform).
6 intersection reduction methods (min, max, product, mean, gmean, softmin) and 
2 difference reduction methods (ignorematch, substractmatch),
2 normalization modes (normalized, and not normalized), and
9 random seeds.
In each case, models were trained for 1000 epochs.
}
\begin{tabular}{lrrrrr}
\toprule
fbank size & n & loss & acc & best acc & p(conv)\\
\midrule
16.0 & 1944 & $0.25 \pm 0.01$ & $0.79 \pm 0.00$ & 1.0 & $0.42 \pm 0.01$\\
8.0 & 1944 & $0.28 \pm 0.01$ & $0.78 \pm 0.00$ & 1.0 & $0.39 \pm 0.01$\\
4.0 & 1944 & $0.30 \pm 0.01$ & $0.76 \pm 0.00$ & 1.0 & $0.38 \pm 0.01$\\
32.0 & 1944 & $0.31 \pm 0.01$ & $0.76 \pm 0.00$ & 1.0 & $0.33 \pm 0.01$\\
2.0 & 1944 & $0.45 \pm 0.01$ & $0.67 \pm 0.00$ & 1.0 & $0.20 \pm 0.01$\\
1.0 & 1944 & $0.54 \pm 0.01$ & $0.61 \pm 0.00$ & 1.0 & $0.12 \pm 0.01$\\
\bottomrule

\end{tabular}
\label{tab:xor-fbank-size-aggregates}
\end{table}

\clearpage
\newpage
\section{Language Modeling with GPT-2 and Tversky-GPT-2 on PTB}
\label{sec:gpt2-ptb-results}
\subsection{Initialization: OpenAI Released Weights}
\begin{table}[ht]
\resizebox{\textwidth}{!}{
\small
\begin{tabular}{llllrrrrrrrrrrrr}
\multicolumn{6}{l}{} & \multicolumn{10}{c}{\textbf{Validation Perplexity}} \\
\cmidrule(lr){7-16}
\multicolumn{3}{l}{} & \multicolumn{3}{r}{difference $\rightarrow$} & \multicolumn{5}{c}{ignorematch} & \multicolumn{5}{c}{substractmatch} \\
\cmidrule(lr){7-16}
\multicolumn{3}{l}{} & \multicolumn{3}{r}{intersection $\rightarrow$} & gmean & max & mean & min & product & gmean & max & mean & min & product \\
\cmidrule(lr){7-11}\cmidrule(lr){12-16}
Init & model & tie-proto & tie-fbank & features & params & PPL & PPL & PPL & PPL & PPL & PPL & PPL & PPL & PPL & PPL \\
\midrule
finetuned & baseline & N &  &  & 163037184 & 34.85 & 34.85 & 34.85 & \textbf{34.85} & 34.85 & \textbf{34.85} & 34.85 & \textbf{34.85} & \textbf{34.85} & \textbf{34.85} \\
\hline
finetuned & tversky-head & N & N & 1024 & 163823619 & 52.42 & 70.11 & 51.80 & 57.66 & 76.73 & 61.59 & 51.90 & 58.15 & 85.37 & 120.22 \\
finetuned & tversky-head & N & N & 2048 & 164610051 & 40.92 & 54.44 & 41.23 & 46.79 & 53.00 & 48.71 & 41.28 & 47.91 & 66.95 & 86.26 \\
finetuned & tversky-head & N & N & 4096 & 166182915 & 34.47 & 38.52 & 34.44 & 40.32 & 40.32 & 42.82 & 34.54 & 41.89 & 64.80 & 57.10 \\
finetuned & tversky-head & N & N & 8192 & 169328643 & \textbf{32.56} & \textbf{32.80} & \textbf{32.53} & 38.80 & 34.50 & 38.89 & \textbf{32.49} & 39.09 & 61.66 & 48.72 \\
finetuned & tversky-head & N & N & 12288 & 172474371 & 32.84 & 32.92 & 32.82 & 40.82 & 32.32 & 41.18 & 32.81 & 41.29 & 70.24 & 52.26 \\
finetuned & tversky-head & N & N & 16384 & 175620099 & 33.62 & 36.33 & 33.62 & 39.75 & $^*$\textbf{32.31} & 40.71 & 33.63 & 40.67 & 77.18 & 52.32 \\
finetuned & tversky-head & N & N & 32768 & 188203011 & 38.36 & 40.34 & 38.98 & 49.35 & 33.94 & 49.48 & 38.79 & 47.17 & 109.10 & 62.37 \\
finetuned & tversky-all-1layer & N & N & 1024 & 114232359 & 145.50 & 208.61 & 142.82 & 154.14 & 219.63 & 172.13 & 144.59 & 155.65 & 638.28 & 297.61 \\
finetuned & tversky-all-1layer & N & N & 2048 & 115018791 & 118.23 & 172.70 & 115.15 & 120.06 & 166.49 & 121.00 & 114.88 & 119.46 & 257.80 & 253.91 \\
finetuned & tversky-all-1layer & N & N & 4096 & 116591655 & 95.65 & 143.67 & 95.51 & 121.13 & 124.50 & 122.75 & 95.79 & 121.51 & 157.91 & 203.54 \\
finetuned & tversky-all-1layer & N & N & 8192 & 119737383 & 91.40 & 94.86 & 91.44 & 93.09 & 99.75 & 93.14 & 91.45 & 93.53 & 149.20 & 142.79 \\
finetuned & tversky-all-1layer & N & N & 12288 & 122883111 & 90.39 & 94.49 & 90.79 & 92.54 & 94.00 & 92.98 & 90.74 & 92.40 & 147.16 & 133.41 \\
finetuned & tversky-all-1layer & N & N & 16384 & 126028839 & 90.51 & 96.41 & 91.22 & 95.39 & 91.08 & 95.64 & 91.23 & 95.06 & 156.30 & - \\
finetuned & tversky-all-1layer & N & N & 32768 & 138611751 & 96.00 & 101.08 & 96.44 & 100.78 & 92.68 & - & - & - & - & - \\
\bottomrule
\end{tabular}}
\caption{Validation perplexity comparison with \textbf{init=finetuned} and \textbf{tie-proto=N}.}
\label{tab:perplexity_init=finetuned_tie_proto=N}
\end{table}
\begin{table}[ht]
\resizebox{\textwidth}{!}{
\small
\begin{tabular}{llllrrrrrrrrrrrr}
\multicolumn{6}{l}{} & \multicolumn{10}{c}{\textbf{Validation Perplexity}} \\
\cmidrule(lr){7-16}
\multicolumn{3}{l}{} & \multicolumn{3}{r}{difference $\rightarrow$} & \multicolumn{5}{c}{ignorematch} & \multicolumn{5}{c}{substractmatch} \\
\cmidrule(lr){7-16}
\multicolumn{3}{l}{} & \multicolumn{3}{r}{intersection $\rightarrow$} & gmean & max & mean & min & product & gmean & max & mean & min & product \\
\cmidrule(lr){7-11}\cmidrule(lr){12-16}
Init & model & tie-proto & tie-fbank & features & params & PPL & PPL & PPL & PPL & PPL & PPL & PPL & PPL & PPL & PPL \\
\midrule
finetuned & baseline & Y &  &  & 124439808 & $^*$\textbf{19.99} & $^*$\textbf{19.99} & $^*$\textbf{19.99} & $^*$\textbf{19.99} & $^*$\textbf{19.99} & $^*$\textbf{19.99} & $^*$\textbf{19.99} & $^*$\textbf{19.99} & $^*$\textbf{19.99} & $^*$\textbf{19.99} \\
\hline
finetuned & tversky-head & Y & N & 1024 & 125226243 & 90.22 & 39.28 & 86.30 & 61.22 & 69.68 & 61.29 & 87.64 & 61.08 & 81.07 & 94.78 \\
finetuned & tversky-head & Y & N & 2048 & 126012675 & 59.95 & 28.69 & 56.18 & 67.96 & 43.19 & 71.83 & 57.38 & 68.19 & 46.20 & 58.08 \\
finetuned & tversky-head & Y & N & 4096 & 127585539 & 23.26 & 27.15 & 24.01 & 24.54 & 28.91 & 23.37 & 24.08 & 23.22 & 30.84 & 35.15 \\
finetuned & tversky-head & Y & N & 8192 & 130731267 & 21.17 & 22.31 & 21.25 & 21.44 & 22.37 & 21.43 & 21.23 & 21.50 & 22.03 & 23.24 \\
finetuned & tversky-head & Y & N & 12288 & 133876995 & 21.07 & 22.64 & 21.09 & 21.33 & 20.99 & 21.32 & 21.08 & 21.35 & 21.98 & 21.81 \\
finetuned & tversky-head & Y & N & 16384 & 137022723 & 20.86 & 21.86 & 20.85 & 21.12 & 20.77 & 21.14 & 20.85 & 21.12 & 22.88 & - \\
finetuned & tversky-head & Y & N & 32768 & 149605635 & 20.70 & 21.33 & 20.68 & 20.82 & 20.46 & - & - & - & - & - \\
finetuned & tversky-all-1layer & Y & N & 1024 & 75634983 & 134.68 & 188.07 & 132.79 & 174.71 & 212.09 & 166.07 & 132.21 & 174.75 & 741.72 & 321.28 \\
finetuned & tversky-all-1layer & Y & N & 2048 & 76421415 & 101.62 & 182.26 & 102.64 & 131.28 & 164.13 & 134.14 & 103.46 & 130.69 & 235.55 & 228.58 \\
finetuned & tversky-all-1layer & Y & N & 4096 & 77994279 & 77.13 & 87.87 & 76.87 & 96.56 & 109.65 & 97.91 & 76.94 & 100.72 & 198.39 & 190.20 \\
finetuned & tversky-all-1layer & Y & N & 8192 & 81140007 & 70.07 & 88.16 & 69.75 & 80.32 & 83.06 & 79.63 & 69.71 & 79.65 & 97.48 & 117.50 \\
finetuned & tversky-all-1layer & Y & N & 12288 & 84285735 & 67.61 & 79.07 & 66.50 & 76.51 & 73.95 & 76.66 & 66.51 & 75.85 & 101.56 & 97.24 \\
finetuned & tversky-all-1layer & Y & N & 16384 & 87431463 & 69.54 & 79.76 & 69.22 & 78.32 & 73.03 & 79.33 & 69.22 & 78.85 & 100.12 & - \\
finetuned & tversky-all-1layer & Y & N & 32768 & 100014375 & 73.33 & 99.82 & 73.12 & 84.67 & 67.19 & - & - & - & - & - \\
\bottomrule
\end{tabular}}
\caption{Validation perplexity comparison with \textbf{init=finetuned} and \textbf{tie-proto=Y}.}
\label{tab:perplexity_init=finetuned_tie_proto=Y}
\end{table}

\clearpage
\newpage

\subsection{Initialization: Random Weights}
\begin{table}[ht]
\resizebox{\textwidth}{!}{
\small
\begin{tabular}{llllrrrrrrrrrrrr}
\multicolumn{6}{l}{} & \multicolumn{10}{c}{\textbf{Validation Perplexity}} \\
\cmidrule(lr){7-16}
\multicolumn{3}{l}{} & \multicolumn{3}{r}{difference $\rightarrow$} & \multicolumn{5}{c}{ignorematch} & \multicolumn{5}{c}{substractmatch} \\
\cmidrule(lr){7-16}
\multicolumn{3}{l}{} & \multicolumn{3}{r}{intersection $\rightarrow$} & gmean & max & mean & min & product & gmean & max & mean & min & product \\
\cmidrule(lr){7-11}\cmidrule(lr){12-16}
Init & model & tie-proto & tie-fbank & features & params & PPL & PPL & PPL & PPL & PPL & PPL & PPL & PPL & PPL & PPL \\
\midrule
scratch & baseline & N &  &  & 163037184 & 134.06 & 134.06 & 134.06 & 134.06 & 134.06 & 134.06 & 134.06 & 134.06 & \textbf{134.06} & 134.06 \\
\hline
scratch & tversky-head & N & N & 1024 & 163823619 & 146.74 & 218.66 & 146.10 & 157.30 & 184.36 & 155.74 & 145.65 & 158.21 & 496.50 & 168.50 \\
scratch & tversky-head & N & N & 2048 & 164610051 & 147.92 & 192.12 & 148.02 & 145.72 & 171.54 & 155.64 & 147.64 & 145.66 & 320.49 & 140.48 \\
scratch & tversky-head & N & N & 4096 & 166182915 & 168.12 & 161.04 & 167.65 & 176.67 & 162.97 & 157.44 & 167.05 & 155.93 & 147.83 & \textbf{133.47} \\
scratch & tversky-head & N & N & 8192 & 169328643 & 210.03 & 204.28 & 217.75 & 182.50 & 173.63 & 182.08 & 215.71 & 182.47 & 205.87 & 154.21 \\
scratch & tversky-head & N & N & 12288 & 172474371 & 227.69 & 178.52 & 241.49 & 212.82 & 186.98 & 212.26 & 236.39 & 212.04 & 203.83 & 163.90 \\
scratch & tversky-head & N & N & 16384 & 175620099 & 229.84 & 202.26 & 250.62 & 228.47 & 213.86 & 231.28 & 248.62 & 231.49 & 228.31 & 200.96 \\
scratch & tversky-head & N & N & 32768 & 188203011 & 182.45 & 162.13 & 190.04 & 190.92 & 252.01 & 198.72 & 187.12 & 197.99 & 223.08 & 218.13 \\
scratch & tversky-all-1layer & N & N & 1024 & 114232359 & nan & 173.25 & 126.67 & 138.92 & 169.06 & nan & 127.19 & 139.09 & 271.09 & 347.79 \\
scratch & tversky-all-1layer & N & N & 2048 & 115018791 & 121.06 & 135.85 & 120.29 & 123.94 & 130.72 & 126.68 & \textbf{120.58} & 124.17 & 163.99 & 249.04 \\
scratch & tversky-all-1layer & N & N & 4096 & 116591655 & \textbf{120.89} & \textbf{127.62} & 121.11 & 123.97 & $^*$\textbf{117.59} & \textbf{124.16} & 121.26 & \textbf{123.86} & 136.27 & 225.91 \\
scratch & tversky-all-1layer & N & N & 8192 & 119737383 & 128.32 & 134.91 & 129.87 & 126.38 & 118.06 & 126.72 & 129.45 & 126.31 & 155.86 & 140.68 \\
scratch & tversky-all-1layer & N & N & 12288 & 122883111 & 130.54 & 131.29 & 132.46 & 127.68 & 125.03 & 127.69 & 132.42 & 127.72 & 159.30 & 137.27 \\
scratch & tversky-all-1layer & N & N & 16384 & 126028839 & 129.24 & 131.03 & 131.82 & 128.37 & 132.29 & 128.33 & 131.94 & 128.01 & 163.26 & - \\
scratch & tversky-all-1layer & N & N & 32768 & 138611751 & 129.11 & 132.33 & 130.26 & 129.51 & 136.46 & - & - & - & - & - \\
\bottomrule
\end{tabular}}
\caption{Validation perplexity comparison with \textbf{init=scratch} and \textbf{tie-proto=N}.}
\label{tab:perplexity_init=scratch_tie_proto=N}
\end{table}
\begin{table}[ht]
\resizebox{\textwidth}{!}{
\small
\begin{tabular}{llllrrrrrrrrrrrr}
\multicolumn{6}{l}{} & \multicolumn{10}{c}{\textbf{Validation Perplexity}} \\
\cmidrule(lr){7-16}
\multicolumn{3}{l}{} & \multicolumn{3}{r}{difference $\rightarrow$} & \multicolumn{5}{c}{ignorematch} & \multicolumn{5}{c}{substractmatch} \\
\cmidrule(lr){7-16}
\multicolumn{3}{l}{} & \multicolumn{3}{r}{intersection $\rightarrow$} & gmean & max & mean & min & product & gmean & max & mean & min & product \\
\cmidrule(lr){7-11}\cmidrule(lr){12-16}
Init & model & tie-proto & tie-fbank & features & params & PPL & PPL & PPL & PPL & PPL & PPL & PPL & PPL & PPL & PPL \\
\midrule
scratch & baseline & Y &  &  & 124439808 & 136.04 & \textbf{136.04} & 136.04 & 136.04 & 136.04 & 136.04 & 136.04 & 136.04 & \textbf{136.04} & \textbf{136.04} \\
\hline
scratch & tversky-head & Y & N & 1024 & 125226243 & 149.22 & 213.67 & 148.81 & 155.98 & 193.72 & 153.13 & 149.07 & 156.29 & 360.26 & 271.20 \\
scratch & tversky-head & Y & N & 2048 & 126012675 & 147.91 & 179.89 & 148.12 & 145.80 & 169.92 & 155.16 & 147.83 & 145.83 & 175.95 & 154.26 \\
scratch & tversky-head & Y & N & 4096 & 127585539 & 167.44 & 168.53 & 168.55 & 150.11 & 161.85 & 151.59 & 168.02 & 149.89 & 152.11 & 145.52 \\
scratch & tversky-head & Y & N & 8192 & 130731267 & 197.06 & 207.89 & 204.79 & 181.21 & 168.29 & 179.76 & 201.95 & 180.99 & 183.96 & 156.37 \\
scratch & tversky-head & Y & N & 12288 & 133876995 & 212.50 & 211.17 & 223.91 & 202.83 & 179.87 & 204.57 & 221.02 & 202.63 & 214.04 & 165.64 \\
scratch & tversky-head & Y & N & 16384 & 137022723 & 210.90 & 152.18 & 227.04 & 217.16 & 204.58 & 214.09 & 226.02 & 216.78 & 238.05 & - \\
scratch & tversky-head & Y & N & 32768 & 149605635 & 162.49 & 151.23 & 171.88 & 176.74 & 216.54 & - & - & - & - & - \\
scratch & tversky-all-1layer & Y & N & 1024 & 75634983 & 142.00 & 174.76 & 138.77 & 152.44 & 182.22 & 154.00 & 138.65 & 152.31 & 250.44 & 389.16 \\
scratch & tversky-all-1layer & Y & N & 2048 & 76421415 & 132.41 & 141.77 & 130.55 & 136.88 & 146.27 & 144.29 & 130.69 & 137.51 & 180.40 & 332.45 \\
scratch & tversky-all-1layer & Y & N & 4096 & 77994279 & \textbf{129.46} & 137.46 & \textbf{129.64} & 130.57 & 128.84 & 130.26 & \textbf{129.66} & 130.61 & 146.93 & 175.75 \\
scratch & tversky-all-1layer & Y & N & 8192 & 81140007 & 136.73 & 145.34 & 139.16 & $^*$\textbf{125.86} & \textbf{128.05} & \textbf{125.98} & 139.13 &\textbf{125.96} & 151.85 & 151.07 \\
scratch & tversky-all-1layer & Y & N & 12288 & 84285735 & 140.86 & 156.21 & 144.50 & 128.38 & 134.98 & 127.85 & 144.61 & 128.45 & 156.44 & 140.26 \\
scratch & tversky-all-1layer & Y & N & 16384 & 87431463 & 143.59 & 160.09 & 148.72 & 130.69 & 140.38 & 129.88 & 148.77 & 130.50 & 153.77 & - \\
scratch & tversky-all-1layer & Y & N & 32768 & 100014375 & 154.05 & - & - & 140.73 & 150.33 & - & - & - & - & - \\
\bottomrule
\end{tabular}}
\caption{Validation perplexity comparison with \textbf{init=scratch} and \textbf{tie-proto=Y}.}
\label{tab:ptb-perplexity_init=scratch_tie_proto=Y}
\end{table}

\clearpage
\newpage
\section{Examples of Semantic Fields in TverskyGPT2's Set-Centric Token Representation}
\label{sec:gpt2-ptb-semantic-fields}
\begin{figure}[h]
    \centering
    \includegraphics[width=0.9\linewidth]{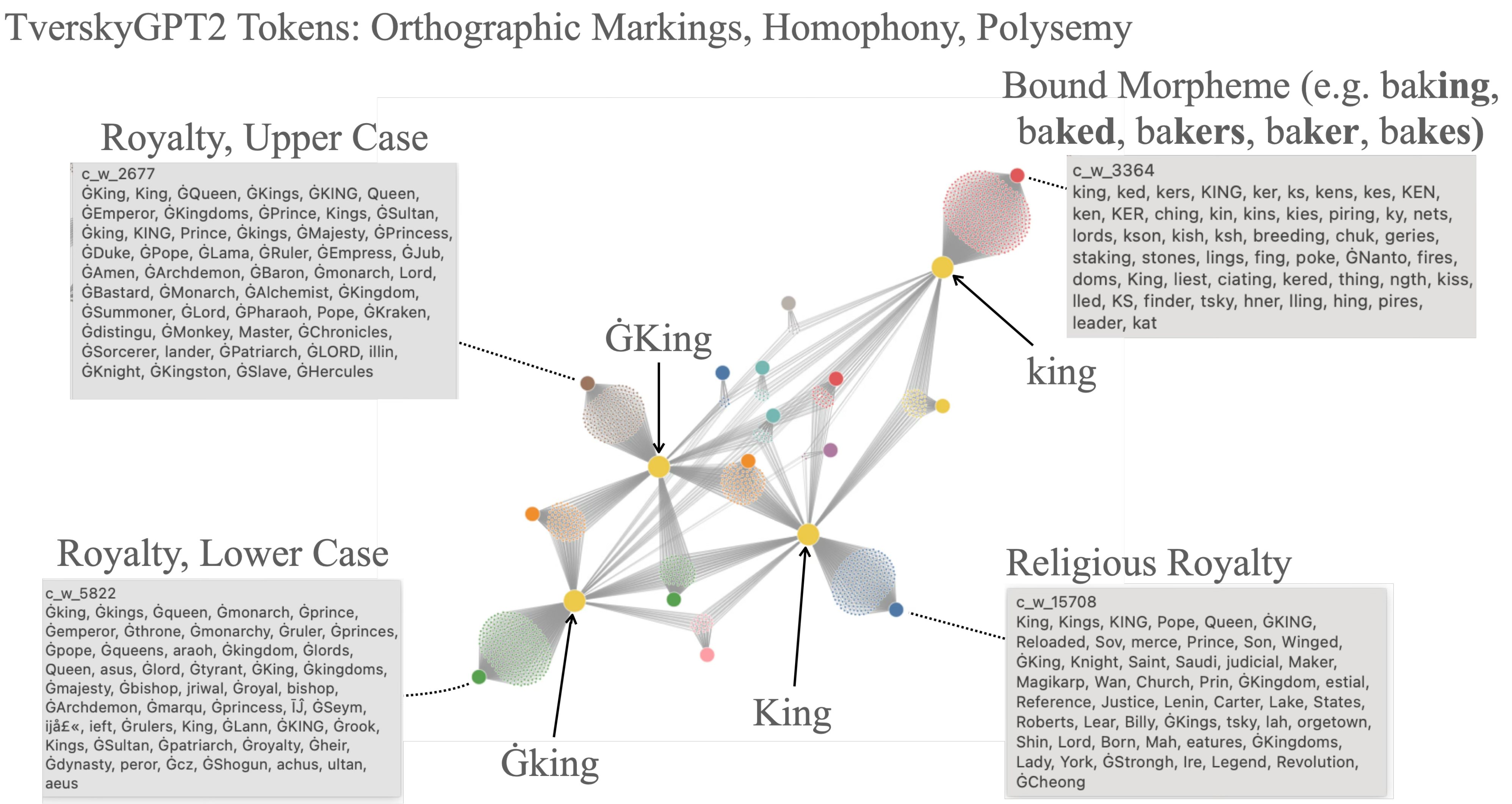}
    \caption{Visualization of the semantic fields formed by the distinctive features of related tokens. Notice that TverskyGPT2 tokens capture semantics related to orthographic markings, and morphological function. The $king$ token (as in baking) is related to other suffixes such as $ked$, $ker$ and $kes$ (as in baked, baker and bakes)}.
    \label{fig:semantic-field-tokens}
\end{figure}

\begin{figure}[h]
    \centering
    \includegraphics[width=0.9\linewidth]{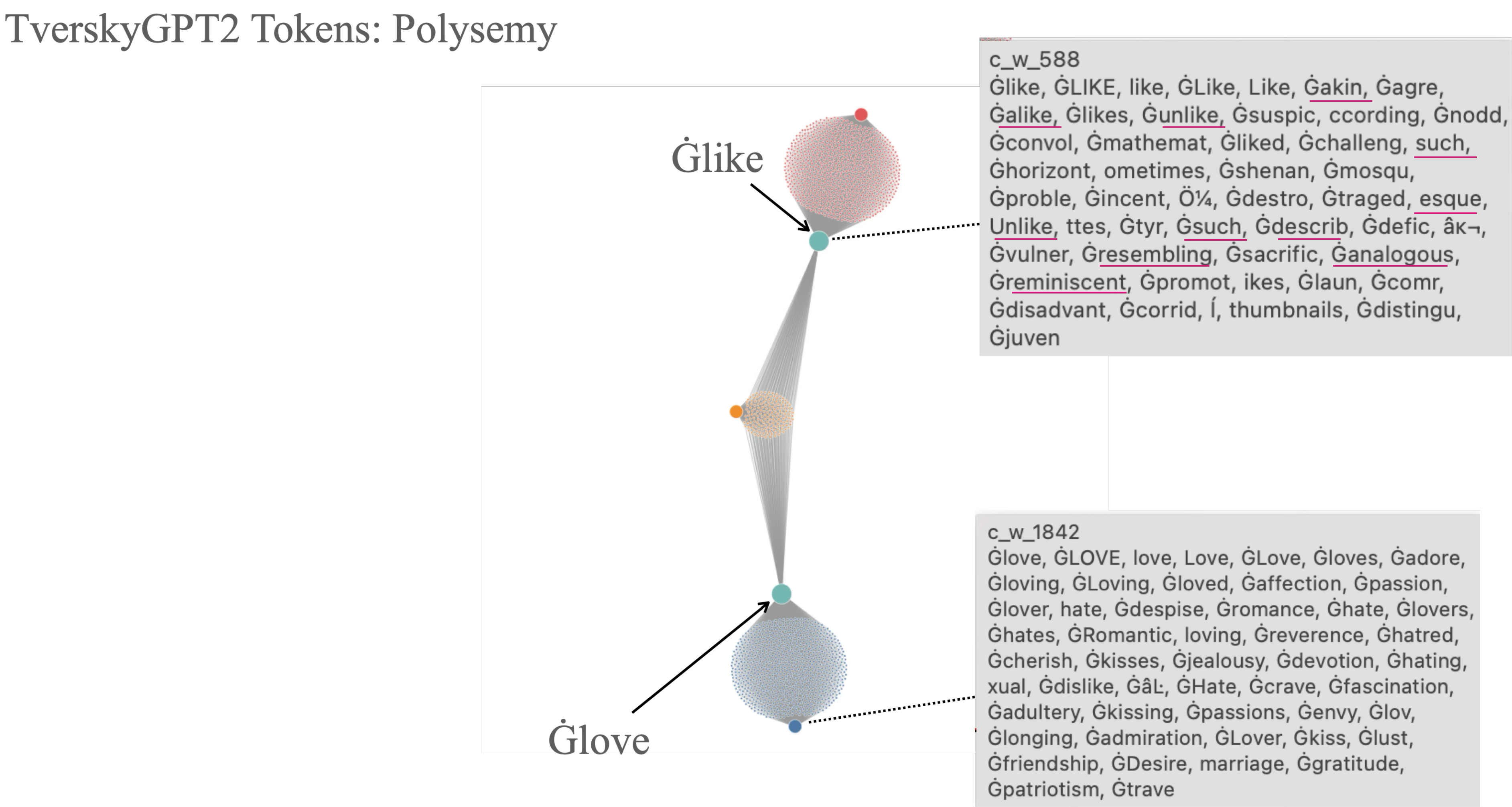}
    \caption{Visualization of the semantic fields formed by the distinctive features of tokens $like$ ($like - love$) and $love$ ($love - like$). $like$, distinctively from $love$, which also captures the sense of sentiment, captures the sense of similarity, with tokens such as \emph{akin}, \emph{alike}, \emph{unlike} and \emph{reminiscent}.}
    \label{fig:semantic-fields-like-love}
\end{figure}

\begin{figure}[h]
    \centering
    \includegraphics[width=0.9\linewidth]{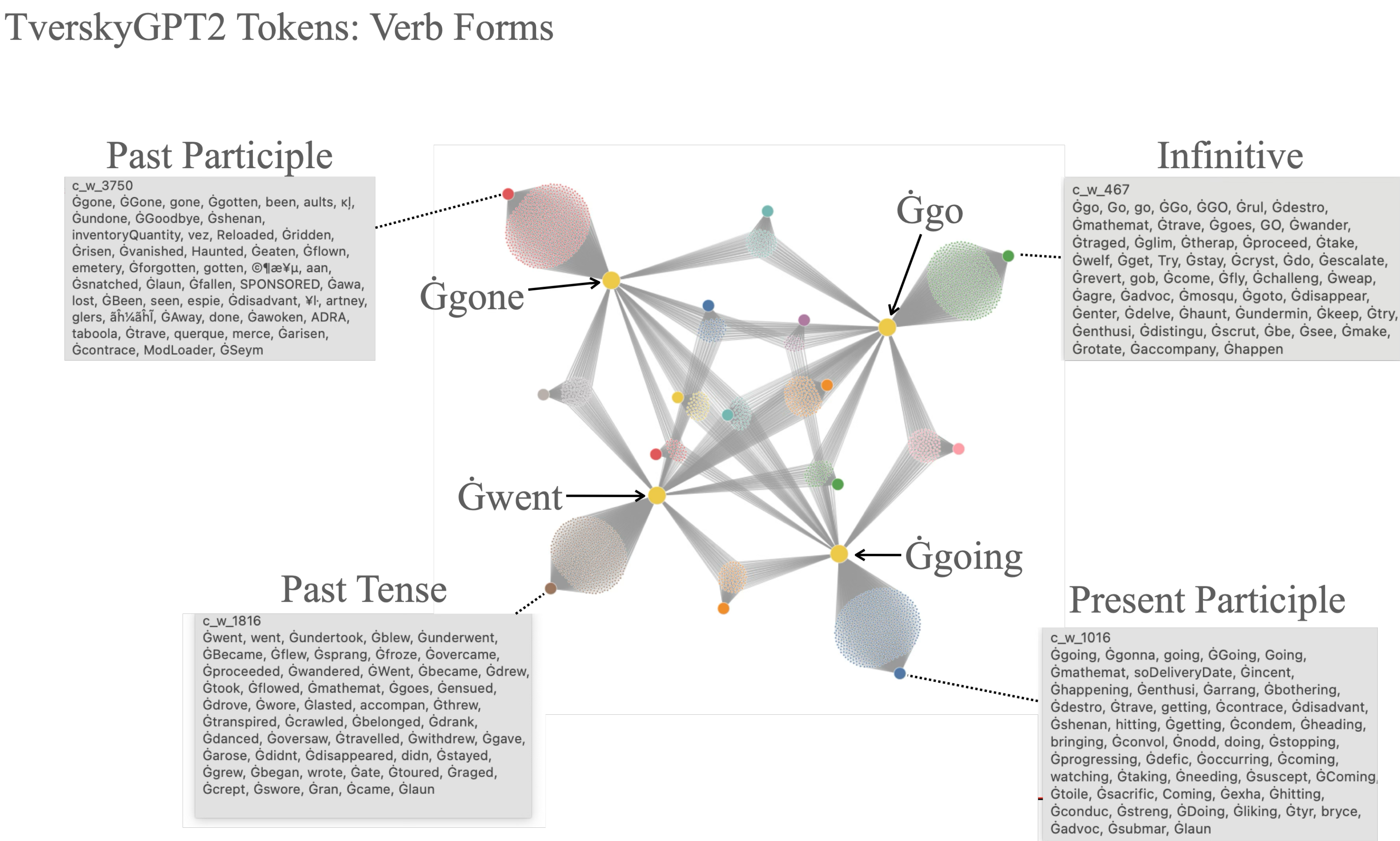}
    \caption{Visualization of the semantic fields formed by the distinctive features of tokens 
    $go$ ($go-gone-went-going$), 
    $gone$ ($gone-go-went-going$), 
    $went$ ($went-gone-go-going$), 
    and $going$ ($going-gone-went-go$), respectively specifying the concept of verb forms \emph{infinitive}, \emph{past participle}, \emph{past tense} and \emph{present participle}.
    }
    \label{fig:semantic-field-go-gone-went-gone}
\end{figure}

\begin{figure}[h]
    \centering
    \includegraphics[width=0.9\linewidth]{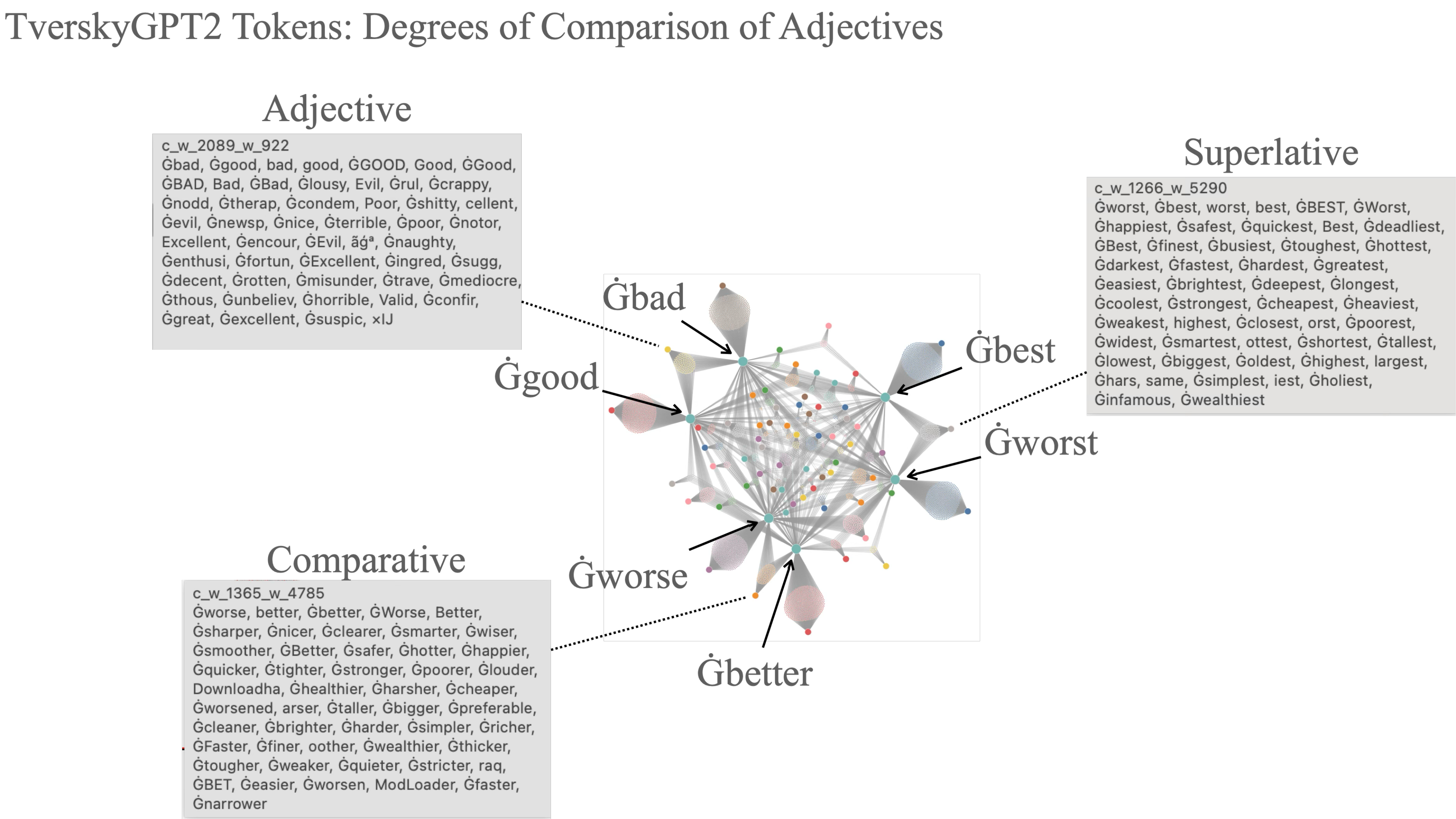}
    \caption{Visualization of the semantic fields formed by the distinctive features of the common features of tokens $bad$ and $good$ ($bad \cap good - worse - better - worst -best$), $worse$ and $better$ ($worse \cap better - bad - good - worst -best$), and $worst$ and $best$ ($worst \cap best - bad - good - worse -better$). These semantic fields algebraically specify the concepts of \emph{adjective}, \emph{comparative}, and \emph{superlative}.}
    \label{fig:semantic-field-adjectives-comparatives-superlatives}
\end{figure}

\begin{figure}[h]
    \centering
    \includegraphics[width=0.9\linewidth]{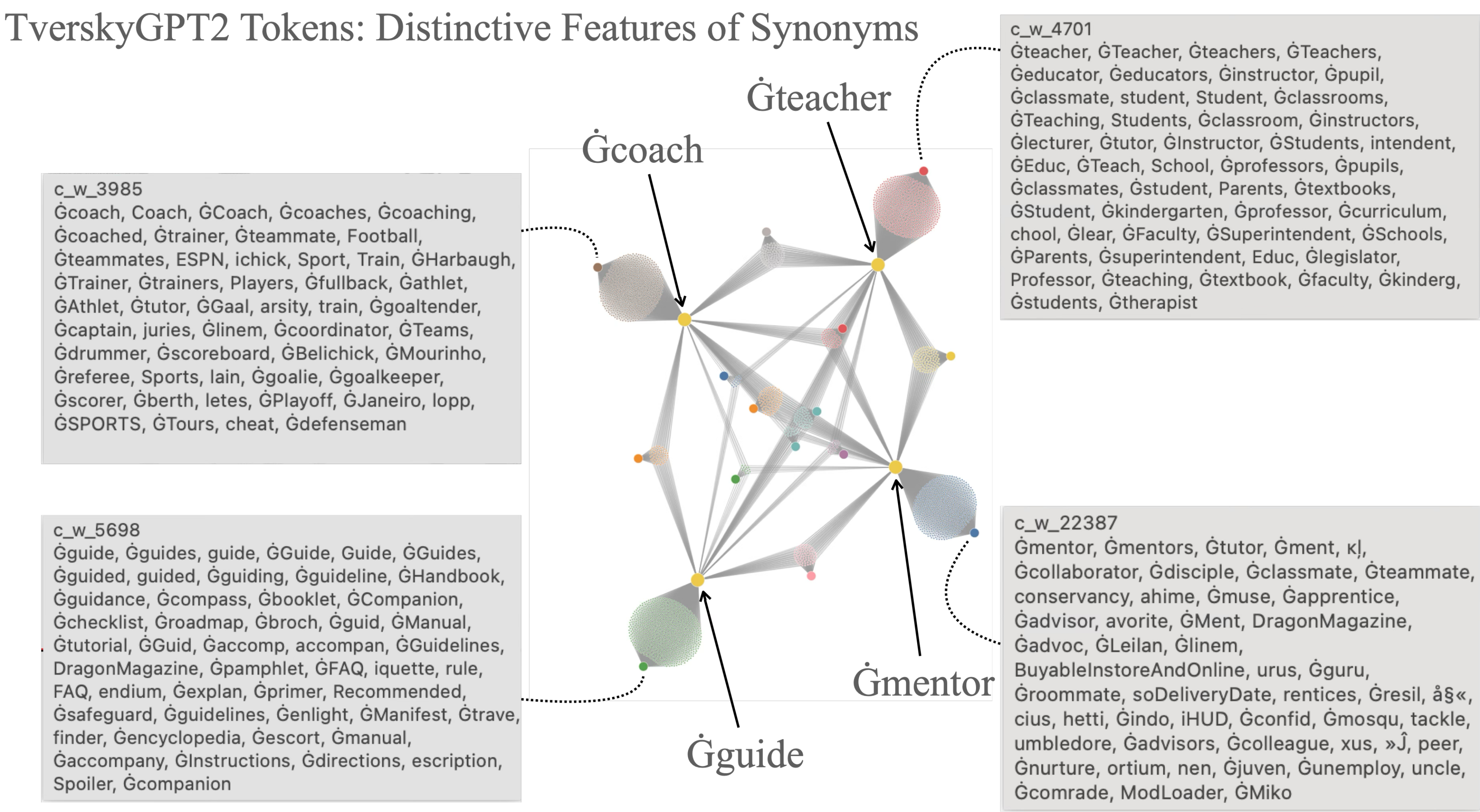}
    \caption{Visualization of the semantic fields capturing the distinctive senses of close synomyms $coach$, $teacher$, $mentor$, and $guide$. Notice that the distinctive features of $coach$ ($coach-teacher-mentor-guide$) shows terms related to sport such as \emph{trainer}, \emph{teammate}, \emph{Football} and \emph{ESPN} whereas the distinctive features of $teacher$ ($teacher -coach-mentor-guide$) show terms related to formal education such as \emph{educator}, \emph{classroom}, \emph{instructor}, \emph{lecturer} and \emph{professor}.}
    \label{fig:semantic-fields-coach-teacher-mentor-guide}
\end{figure}

Figures
~\ref{fig:semantic-field-tokens}
,\ref{fig:semantic-fields-like-love}
,\ref{fig:semantic-field-go-gone-went-gone}
,\ref{fig:semantic-field-adjectives-comparatives-superlatives}
 and \ref{fig:semantic-fields-coach-teacher-mentor-guide} show examples of visualizations of semantic fields formed using set expressions on TverskyGPT2 tokens represented as sets. 
 For each token, the top 1000 features (sorted by dot-product with the token) are considered, and represented as the smallest colored circles.
 Tokens are represented as yellow circles, and connected to the features they comprise with grey lines.
 All distinct intersections of features are colored in the same color, and connected to a circle of the same color representing a distinctive semantic field. Semantic fields of interest are annotated with a text box listing the top 50 tokens by semantic score (See Section~\ref{sec:results:qualitative-analysis}) in that semantic field, permitting its visualization.
 A force-directed graph layout is employed to position the circles corresponding to tokens, features, and semantic fields.
 Note that the \.{G} character (U+0120) represents space in GPT2 tokenization, distinguishing tokens that appear at the beginning of words from other tokens.

\begin{figure}[h]
    \centering
    \includegraphics[width=0.9\linewidth]{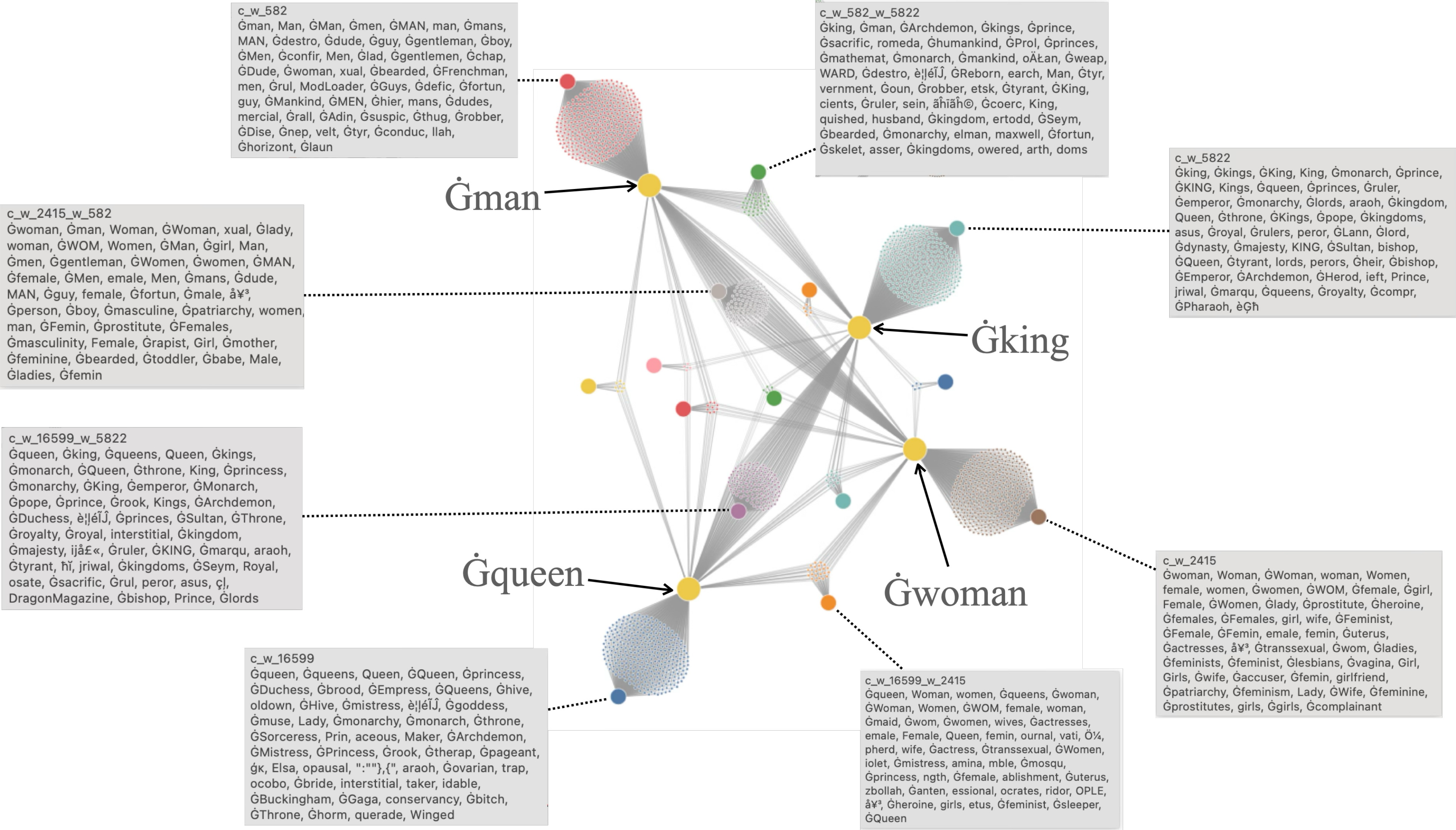}
    \caption{Visualization of the semantic fields capturing the distinctive feature of tokens $man$, $woman$, $king$ and $queen$, and the distinctive semantics of the common features of  $man$ and $king$, $man$ and $woman$, $woman$ and $queen$, $queen$ and $king$, and  $man$ and $woman$}
    \label{fig:semantic-fields-man-woman-king-queen}
\end{figure}

% removed becaused perplexities are too high. needed to be trained for longer, or on more data. prioritizing wikitext-103
%\clearpage
%\newpage
%\section{Tversky-GPT-2 on Wikitext-2}
%\input{exp_006/tables_wikitext2}

%\clearpage
%\newpage
%\subsection{Tversky-GPT-2 on Wikitext-103}
%\input{exp_006/tables_wikitext_103}

% TODO: review this
%\clearpage
%\newpage
%\input{appendix_resnet50_mnist}

\clearpage
\newpage
\section{[Visual]MNISTNet, and [Visual]TverskyMNISTNet, }
\label{sec:tversky-mnistnet}
\begin{figure}[h]
    \centering
    \includegraphics[width=0.7\linewidth]{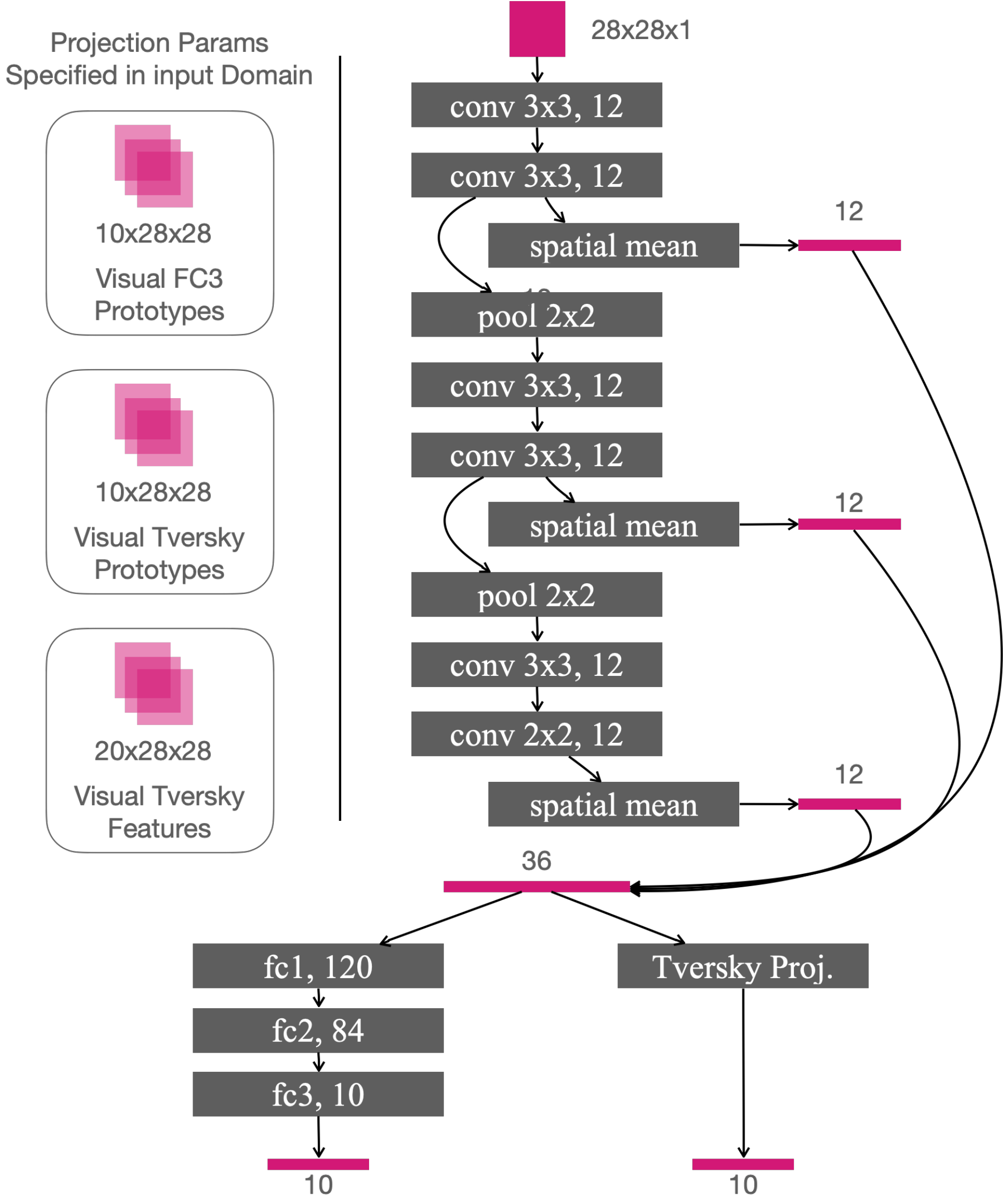}
    \caption{The neural network architecture used in our qualitative analysis experiments.}
    \label{fig:mnist-cnns}
\end{figure}

Two neural networks, VisualMNISTNet, and VisualTverskyMNISTNet (Figure~\ref{fig:mnist-cnns}) are trained to perform MNIST handwritten digit classification. The two neural network architectures share the same convolutional feature extraction stack yielding a 36-dimensional vector representation of the input image. They differ in  how those vectors are projected onto the 10-dimensional output vectors corresponding to 10 digit classes. VisualMNISTNet employs a stack of 120, 84, and 10 unit linear projection layers following \citet{lecun1998gradient}. However, the ReLU activation function is used instead of the logistic sigmoid. VisualTverskyMNISTNet employs a single Tversky Projection layer with 10 prototypes and 20 features. Both neural networks employ the method described in Section~\ref{sec:method:data-domain-param-visualization}, facillitating our qualitative analysis of the learned prototypes and features.

Two additional neural networks, MNISTNet and TverskyMNISTNet, which are identical to their \emph{Visual} counterparts, but do not employ our parameter visualization method are also trained to serve as baseline for accuracy. These neural networks have fewer parameters as the projection parameters are specified in their vector form, which is more compact. 

MNISTNet 3 times smaller than LeNet-5 due to the  design of its convolution stack, which outputs a 36-dimensional vector which are the concatenation of three 12-dimensional vectors obtained by averaging 3 convolutional feature maps across their spatial dimensions.

TverskyMNISTNet, with only 7K parameters is also 3 times smaller than MNISTNet because it employs a single Tversky Projection layer instead of a stack linear projection layers and non-linear activation functions.
All 4 neural networks are trained for 1000 epochs with a batch size of 500. Dropout is applied to the output of the convolution stack at the rate of 0.05. Adam optimizer is used with learning rate 0.002, and weight decay rate 0.00001.

\begin{table}[h]
    \centering
    \caption{Parameter count and accuracy
    of baseline and tversky convolutional neural networks trained for MNIST handwritten digit classification. The "visual" variants specify tversky prototypes and features, and the third fully connected layer's parameters in the input space as 28x28 matrices. All models were trained for 1000 epochs.}
    \begin{tabular}{lrr}
        \toprule
        Model & Params & Valid ACC  \\
        \midrule        
        MNISTNet & 21\,394 & 99.1\% \\
        VisualMNISTNet & 28\,394 & 99.1\% \\
        \midrule
        TverskyMNISTNet & 7\,023 & 98.7\% \\
        VisualTverskyMNISTNet  & 29\,463 & 98.7\% \\
        \bottomrule
    \end{tabular}
    \label{tab:mnist-interpretability-experiments}
\end{table}

\begin{figure}
    \centering
    \includegraphics[width=0.95\linewidth]{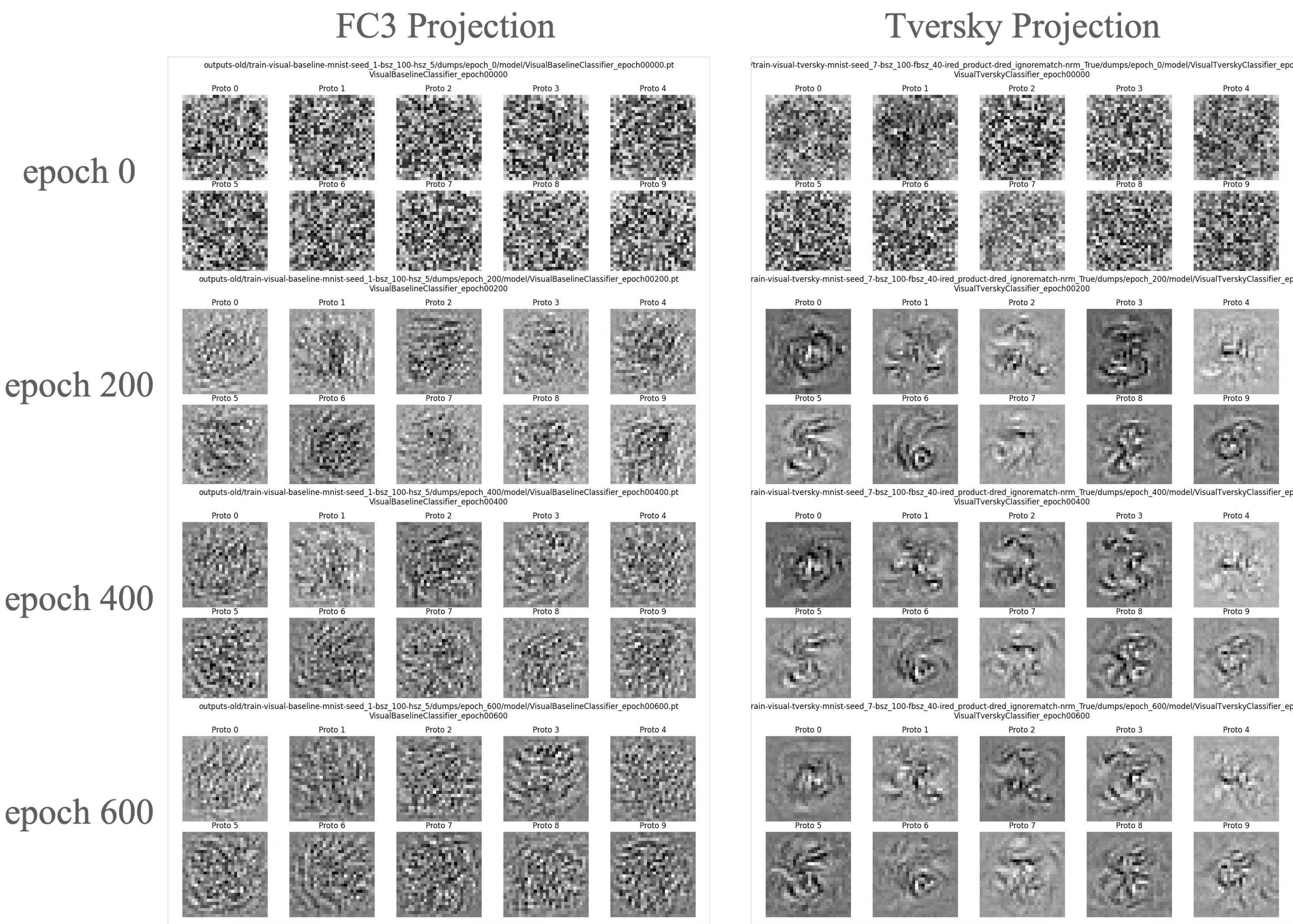}
    \caption{Learned visual prototypes in the baseline and tversky variants of the neural network described in Figure~\ref{fig:mnist-cnns} after epochs 0, 200, 400 and 600.}
    \label{fig:mnist-net-prototypes-over-iterations}
\end{figure}

\begin{figure}[h]
    % \centering\includegraphics[width=0.5\linewidth]{method_illustrations/mnist_features_venn_diagram.pdf}
    \centering\includegraphics[width=0.9\linewidth]{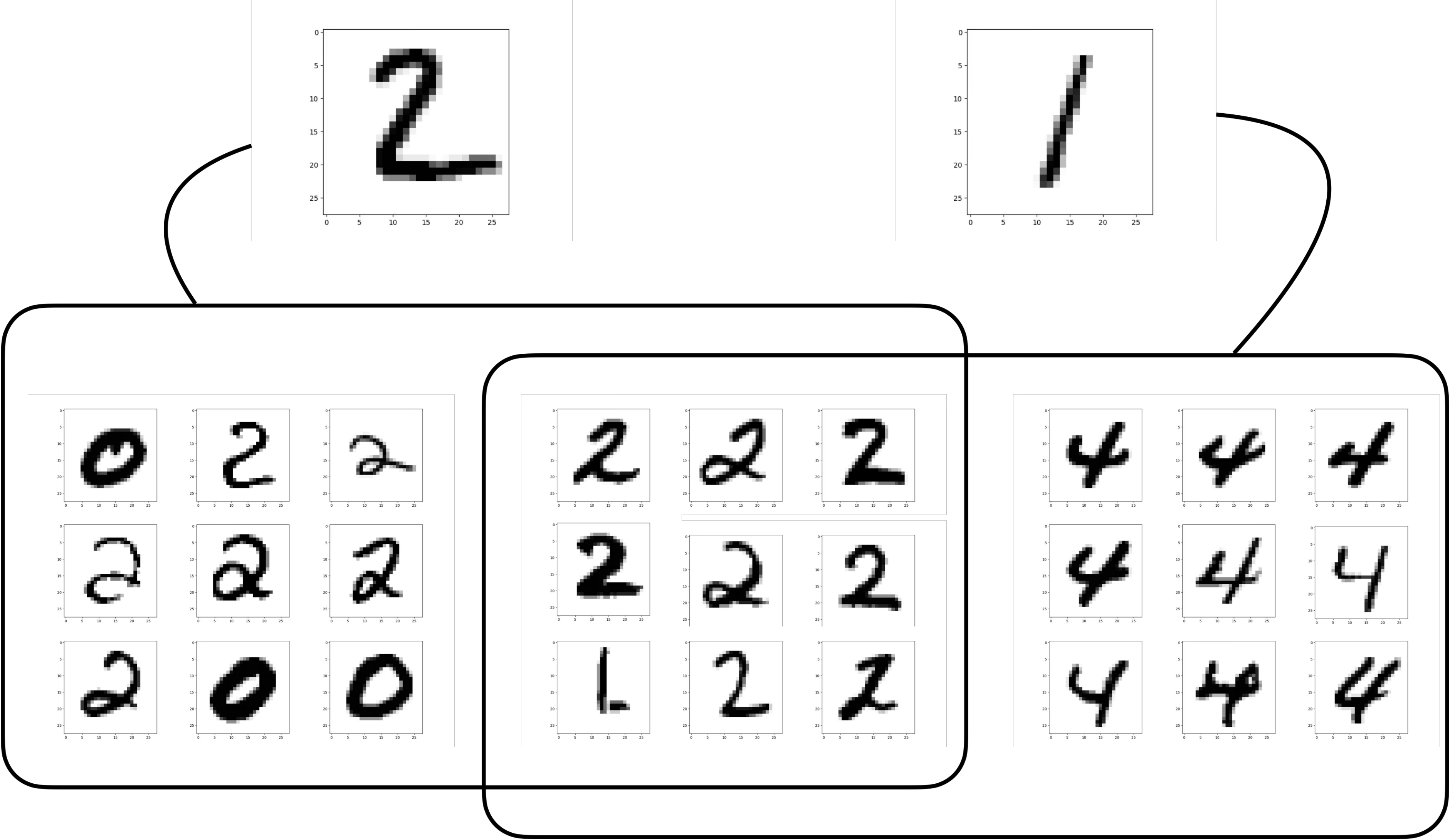}
    \caption{Two MNIST digits: a two (A), and a one (B). A grid of top-9 instances ranked by semantic score in the $A-B$, $A \cap B$ and $B-A$ semantic fields are shown (see Section~\ref{sec:results:qualitative-analysis}). The distinctive features of A exhibit curviness. The distinctive features of B exhibit a vertical stroke slanted to the right.}
    \label{fig:tversky-mnist-venn}
\end{figure}

\clearpage
\newpage

\section[Example of learned alpha, beta, and theta]{Example of learned $\alpha$, $\beta$ and $\theta$}
\label{sec:example-of-learned-alpha-beta-theta}

\begin{figure}[h]
    \centering
    \includegraphics[width=0.9\linewidth]{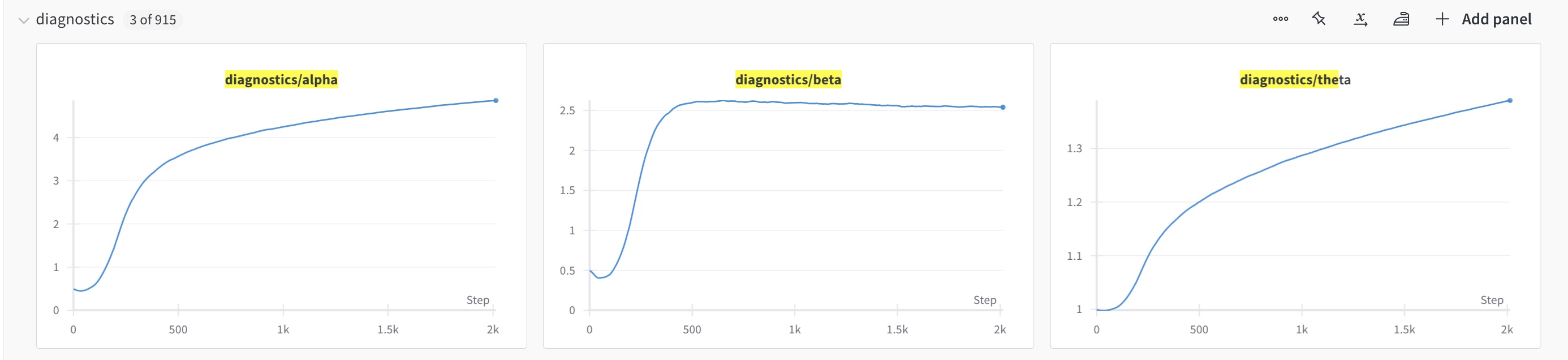}
    \caption{Example of learned $\alpha$, $\beta$ and $\theta$ values over the training iterations of a Tversky-Resnet-50 on the NABirds dataset. As per Tversky's hypothesis of prototypicality, $\alpha$ > $\beta$ in the trained model. The distinctive features of instances are weighted more than the distinctive features of prototypes.}
    \label{fig:resnet-50-nabirds-learned-alpha-beta-theta}
\end{figure}

\clearpage
\newpage
\section{Computation Times and Resources}
\label{sec:computation-times-and-resources}

\subsection{Resnet-50 / NABirds Experiments}

\begin{table}[h]
    \centering
    \caption{Device use and wall-clock times for the NABirds experiments}
    \begin{tabular}{lll}
        \toprule
        Model & Time & Devices \\
        \midrule
         Resnet-50 & 2h 43m 4s &  4 x NVIDIA RTX A6000\\
         Resnet-50 & 3h 34m 37s &  4 x NVIDIA RTX A6000\\
         Resnet-50 & 2h 45m 47s &  4 x NVIDIA RTX A6000\\
         Resnet-50 & 3h 25m 7s &  4 x NVIDIA RTX A6000\\
         \midrule
         Tversky-Resnet-50 & 2h 43m 17s & 4 x NVIDIA RTX 6000 Ada Generation \\
         Tversky-Resnet-50 & 2h 58m 29s & 4 x NVIDIA RTX 6000 Ada Generation \\
         Tversky-Resnet-50 & 2h 41m 46s & 4 x NVIDIA RTX 6000 Ada Generation \\
         Tversky-Resnet-50 & 2h 57m 21s & 4 x NVIDIA RTX 6000 Ada Generation \\
         \bottomrule
    \end{tabular}
    \label{tab:exp-001.02-time-and-resources}
\end{table}

\iffalse
\subsection{Resnet-50 / MNIST Experiments}

\begin{figure}[h]
    \centering
    \includegraphics[width=0.6\linewidth]{exp_003/mnist_experiment_wallclock_times.png}
    \caption{Wall-clock Times in seconds for [Tversky-]Resnet-50 experiments on MNIST. 4 x NVIDIA RTX A6000 were used for each run. The x-axis shows Tversky-Resnet-50 feature bank sizes. Baseline Resnet-50 are ploted at x=0.}
    \label{fig:exp-003.01-time-and-resources}
\end{figure}

\fi

\subsection{GPT2 Experiments}

\begin{figure}[h]
    \centering
    \includegraphics[width=0.6\linewidth]{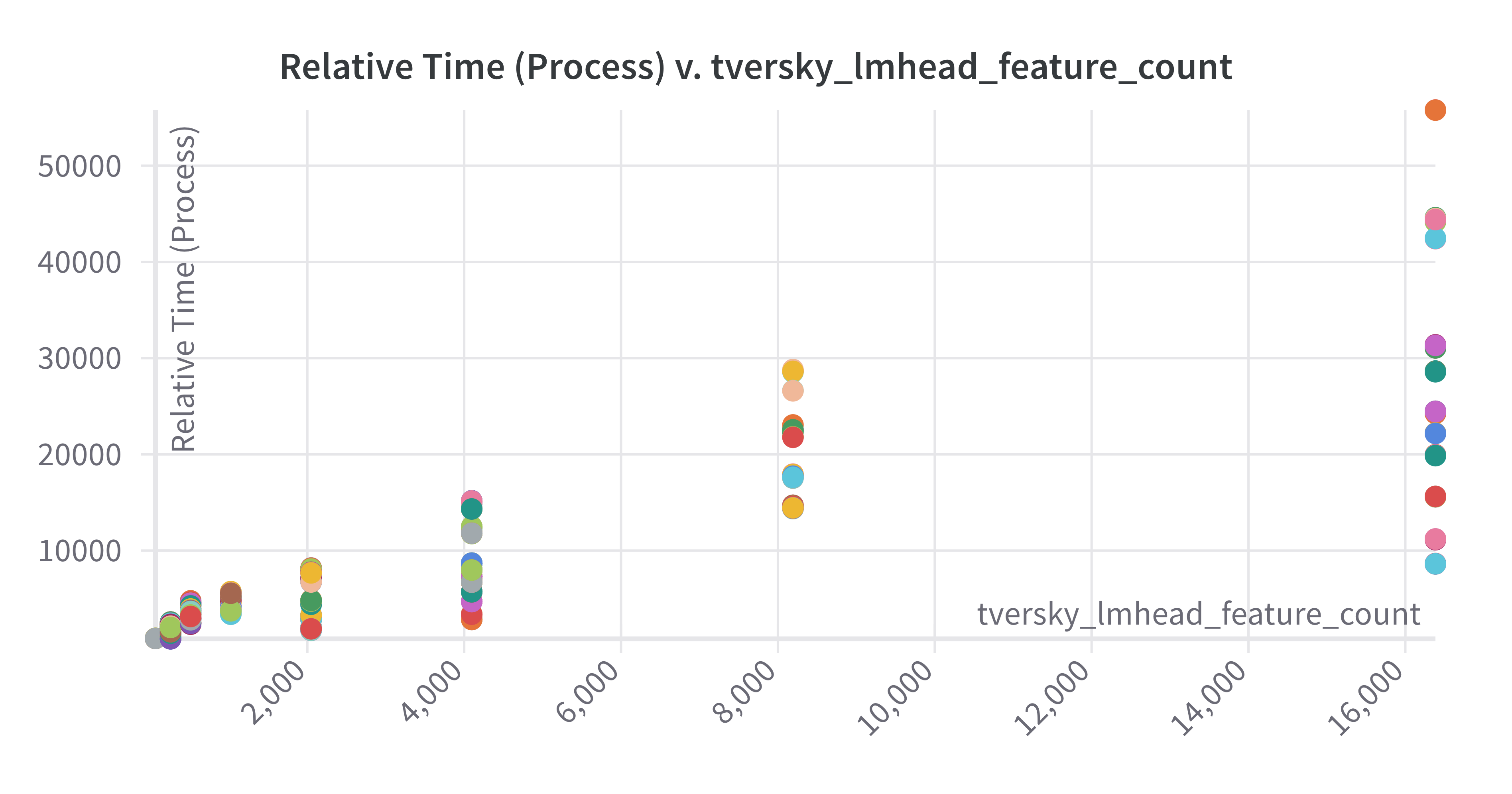}
    \caption{Sample of wall-clock times in seconds for TverskyGPT2 experiments on PTB as a function of the feature bank size. 2 GPUs were used for each run. GPU models varied. They included NVIDIA A100-SXM4-40GB, NVIDIA H100 80GB HBM3, and NVIDIA RTX 6000 Ada Generation.}
    \label{fig:exp-006-time-and-resources}
\end{figure}

%\clearpage
%\newpage
%\input{todo}

%%%%%%%%%%%%%%%%%%%%%%%%%%%%%%%%%%%%%%%%%%%%%%%%%%%%%%%%%%%%%%%%%%%%%%%%%%%%%%%
%%%%%%%%%%%%%%%%%%%%%%%%%%%%%%%%%%%%%%%%%%%%%%%%%%%%%%%%%%%%%%%%%%%%%%%%%%%%%%%

%%%%%%%%%%%%%%%%%%%%%%%%%%%%%%%%%%%%%%%%%%%%%%%%%%%%%%%%%%%%%%%%%%%%%%%%%%%%%%%
%%%%%%%%%%%%%%%%%%%%%%%%%%%%%%%%%%%%%%%%%%%%%%%%%%%%%%%%%%%%%%%%%%%%%%%%%%%%%%%
% NEURIPS CHECKLIST
%%%%%%%%%%%%%%%%%%%%%%%%%%%%%%%%%%%%%%%%%%%%%%%%%%%%%%%%%%%%%%%%%%%%%%%%%%%%%%%
%%%%%%%%%%%%%%%%%%%%%%%%%%%%%%%%%%%%%%%%%%%%%%%%%%%%%%%%%%%%%%%%%%%%%%%%%%%%%%%

% \newpage
% \clearpage
% \input{sec_neurips_checklist}

\end{document}